%% file: main.tex
\documentclass[dvipsnames]{article}

\usepackage[final]{neurips_2021}
\usepackage[numbers]{natbib}

\usepackage[utf8]{inputenc} 
\usepackage[T1]{fontenc}    
\usepackage{url}            
\usepackage{booktabs}       
\usepackage{amsfonts}       
\usepackage{nicefrac}       
\usepackage{booktabs} 
\usepackage{multirow}
\usepackage{wrapfig}
\usepackage{amssymb}
\usepackage{epigraph}
\usepackage{makecell}
\usepackage[edges]{forest}
\usepackage{xcolor}     
\usepackage[labelfont=bf]{caption} 
\usepackage[ruled,vlined]{algorithm2e}
\usepackage{color, soul}
\usepackage{microtype}      
\usepackage{xcolor}         
\usepackage{geometry}
\usepackage{times}
\usepackage{ragged2e}
\usepackage{amsmath}
\usepackage{bm}
\usepackage{latexsym}
\usepackage{subfigure}
\usepackage{graphicx}
\usepackage{float}
\usepackage{longtable}
\usepackage{booktabs} 
\usepackage{multirow}
\usepackage{amssymb}
\usepackage{longtable}
\usepackage{tabu}
\usepackage{bbding}
\usepackage{awesomebox} 
\usepackage{bbding}
\usepackage[most]{tcolorbox}
\usepackage{etoolbox}
\usepackage{fancyhdr}
\usepackage{nicematrix}
\usepackage{cite}

\usepackage{caption} 
\usepackage{chngcntr}

\usepackage[colorlinks, linkcolor=RoyalBlue, anchorcolor=BrickRed, citecolor=RoyalBlue, urlcolor=RoyalBlue]{hyperref}

\newcommand{\ybsun}[1]{{\color{black}#1}} 

\NewDocumentCommand{\ning}
{ mO{} }{\textcolor{purple}{\textsuperscript{\textit{ning}}\textsf{\textbf{\small[#1]}}}}

\usepackage{cleveref}
\crefname{section}{§}{§§}
\Crefname{section}{§}{§§}

\newcommand\refsec[1]{\hyperref[sec:#1]{§\ref{sec:#1}:~\textsc{#1}}}
\newcommand\refsecs[2]{\hyperref[sec:#1]{§\ref{sec:#1}:~\textsc{#1}}, \hyperref[sec:#2]{§\ref{sec:#2}:~\textsc{#2}}}

\newtcolorbox{myboxnote}[1][]{
  breakable,
  title=#1,
  colback=cyan!0,
  colbacktitle=cyan!0,
  coltitle=black,
  fonttitle=\bfseries,
  bottomrule=0pt,
  toprule=0pt,
  leftrule=1.5pt,
  rightrule=1.5pt,
  titlerule=0pt,
  arc=0pt,
  outer arc=0pt,
  colframe=lightgray,
}
\newenvironment{itemize*}%
 {\leftmargini=20pt\begin{itemize}%
  \setlength{\itemsep}{3pt}%
  \setlength{\parskip}{0pt}%
  }%
 {\end{itemize}}
\newenvironment{enumerate*}%
 {\begin{enumerate}%
  \setlength{\itemsep}{0pt}%
  \setlength{\parskip}{0pt}}%
 {\end{enumerate}}

\usepackage{fancyhdr} 
\usepackage{blindtext} 

\tikzset{%
    parent/.style =          {align=center,text width=2cm,rounded corners=3pt, line width=0.3mm, fill=gray!10,draw=gray!80},
    child/.style =           {align=center,text width=2.3cm,rounded corners=3pt, fill=blue!10,draw=blue!80,line width=0.3mm},
    grandchild/.style =      {align=center,text width=2cm,rounded corners=3pt},
    greatgrandchild/.style = {align=center,text width=1.5cm,rounded corners=3pt},
    greatgrandchild2/.style = {align=center,text width=1.5cm,rounded corners=3pt},    
    referenceblock/.style =  {align=center,text width=1.5cm,rounded corners=2pt},
    o_ssm/.style =           {align=center,text width=1.8cm,rounded corners=3pt, fill= magenta!10,draw= magenta!80,line width=0.3mm},   
    o_ssm_work/.style =           {align=center, text width=8cm,rounded corners=3pt, fill= magenta!10,draw= magenta!0,line width=0.3mm},  
    s4/.style =           {align=center,text width=1.8cm,rounded corners=3pt, fill=blue!10,draw=blue!80,line width=0.3mm},   
    s4_work/.style =           {align=center,text width=8cm,rounded corners=3pt, fill=blue!10,draw=blue!0,line width=0.3mm},    
    mamba/.style =           {align=center,text width=1.8cm,rounded corners=3pt, fill= orange!10,draw= orange!80,line width=0.3mm},   
    mamba_work/.style =           {align=center,text width=5.5cm,rounded corners=3pt, fill= orange!10,draw= orange!0,line width=0.3mm},    
    mamba_sub_work/.style = {align=center,text width=5.5cm,rounded corners=3pt, fill= orange!10,draw= orange!0,line width=0.3mm},
    mamba_app/.style =           {align=center,text width=8cm,rounded corners=3pt, fill= orange!10,draw= orange!0,line width=0.3mm},
}

\title{Technologies on Effectiveness and Efficiency: A Survey of State Spaces Models}

\author{%
   Xingtai Lv$^{1}$\thanks{equal contributions}, Youbang Sun$^{1*}$, Kaiyan Zhang$^{1*}$, Shang Qu$^{1,2}$, Xuekai Zhu$^{1}$ \\\textbf{Yuchen Fan$^{1,2}$, Yi Wu$^{3}$, Ermo Hua$^{1}$, Xinwei Long$^{1}$, Ning Ding$^{1,2}$\thanks{corresponding authors}, Bowen Zhou$^{1,2\dag}$}
   \\ $^{1}$Department of Electronic Engineering, Tsinghua University
   \\$^{2}$Shanghai Artificial Intelligence Laboratory, $^{3}$Robotics Institute, Carnegie Mellon University
   \\ \texttt{lvxt24@mails.tsinghua.edu.cn}
   }

\begin{document}

\maketitle

\begin{abstract}

State Space Models (SSMs) have emerged as a promising alternative to the popular transformer-based models and have been increasingly gaining attention. 
Compared to transformers, SSMs excel at tasks with sequential data or longer contexts, demonstrating comparable performances with significant efficiency gains.
In this survey, we provide a coherent and systematic overview for SSMs, including their theoretical motivations, mathematical formulations, comparison with existing model classes, and various applications.
We divide the SSM series into three main sections, providing a detailed introduction to the original SSM, the structured SSM represented by S4, and the selective SSM typified by Mamba.
We put an emphasis on technicality, and highlight the various key techniques introduced to address the effectiveness and efficiency of SSMs.
We hope this manuscript serves as an introduction for researchers to explore the theoretical foundations of SSMs.

\end{abstract}


\section{Introduction}
\input{sections/1_introduction}

\section{State Space Models: From Continuous To Discrete}
\label{State Space Models: From Continuous To Discrete}
\input{sections/2_ssm}

\section{Structured {State Space Models}}
\label{S4: More Effective and Efficient}
\input{sections/3_s4}

\section{Selective {State Space Models}}
\label{Mamba: More Practical and Powerful}
\input{sections/4_mamba}

\section{{Relationships with} Other Model Architectures}
\label{SSM and Other Model Architecture}
\input{sections/5_models}

\section{Applications}
\label{Application}
\input{sections/10_application}

\section{Conclusion}

This survey provides an overview of the theoretical motivations, evolutionary trajectory, comparisons with existing models, and real-world applications of the State Space Models (SSMs) series. 
The SSM series has evolved from its initial conception into advanced architectures such as S4 and Mamba, progressing through stages from the original SSM to structured SSM and selective SSM. 
This transformation has been driven by techniques aimed at enhancing model effectiveness and computational efficiency. These techniques form the core focus of this survey.
SSMs have found extensive applications in domains that require sequence understanding and prediction, such as natural language processing, video comprehension, and time series analysis. 
With the integration of increasingly sophisticated techniques to improve both effectiveness and efficiency, SSMs are expected to play a significant role in tackling more complex challenges.

\newpage
\bibliographystyle{unsrt}
\bibliography{custom,application}


\end{document}

%% file: sections/1_introduction.tex
Large language models are playing an increasingly significant role in various aspects of real-world applications. Although the Transformer architecture~\citep{vaswani2017attention} remains the dominant framework for mainstream language models, alternative architectures have emerged, aiming to address some of the inherent limitations of Transformers. Among these non-Transformer architectures, the State Space Model (SSM), particularly Mamba~\citep{gu2023mamba} and its variants, has attracted considerable attention and gained widespread application.
Compared to \ybsun{Transformers, state space model-based algorithms}
exhibit immense potential for computational efficiency, excelling in handling long-text tasks. 
\ybsun{Moreover, }the SSM series of models have undergone extensive development and refinement, enabling them to handle various data formats that can be serialized, such as text, images, audio, and video. 
\ybsun{Their performance on these practical tasks often rivals that of Transformers. Consequently, these models}
have found widespread use in domains such as healthcare, conversational assistants, and the film industry~\citep{wang2024mambaunetunetlikepurevisual, yang2024vivimvideovisionmamba, guo2024mambamorphmambabasedframeworkmedical, li20243dmambacompleteexploringstructuredstate, ma2024umambaenhancinglongrangedependency, xing2024segmambalongrangesequentialmodeling, elgazzar2022fmris4learningshortlongrange, du2024spikingstructuredstatespace}.

The development of the SSM series can be broadly categorized into three distinct stages, marked by three milestone models: the original SSM, the Structured State Space Sequence Model (S4)~\citep{gu2021efficiently}, and Mamba~\citep{gu2023mamba}.
Initially, the SSM \ybsun{formulation} was proposed to describe \ybsun{physical systems in continuous time.}
\ybsun{The SSMs are then discretized for computer analysis with tractable computations. A}fter discretization, the SSMs can model sequence data, which marks the first stage of the SSM's development. 
However, these early SSMs were rudimentary and faced several limitations, including weak data-fitting capabilities, lack of parallel computation, and susceptibility to overfitting, making them inadequate for practical applications. 
Despite these shortcomings, these models offered advantages such as low computational complexity, flexible core formulas that allowed for derivation into various forms, and significant potential for future improvements. 
S4 and its variant models represent the second stage in the development of the SSM series. S4 took into account the time-invariant situation and utilized the convolutional expression form of the core formula of SSM, dramatically reducing computational complexity and enabling efficient computation. This advancement resolved challenges such as exponential memory decay and the inability to capture long-range dependencies effectively, significantly improving the utility of SSMs. Building on the S4 architecture, optimized models like DSS~\citep{gupta2022diagonalstatespaceseffective} and S4D~\citep{gu2022parameterizationinitializationdiagonalstate} were introduced. Nevertheless, S4 hardly had any optimizations for the underlying computing hardware and usually struggled to achieve satisfactory performance in practical tasks.
The third stage began with the advent of Mamba. By introducing selectivity, optimizing the underlying hardware, and improving the calculation formulas, Mamba achieved a better adaptation to current computing hardware and attained practical task performances close to those of the Transformer. Based on Mamba, numerous model architectures developed for practical application scenarios have been continuously proposed, and models that combine the Mamba architecture with the Transformer architecture are also being explored.

\begin{figure}[h]
\centering
\includegraphics[width=1\textwidth]{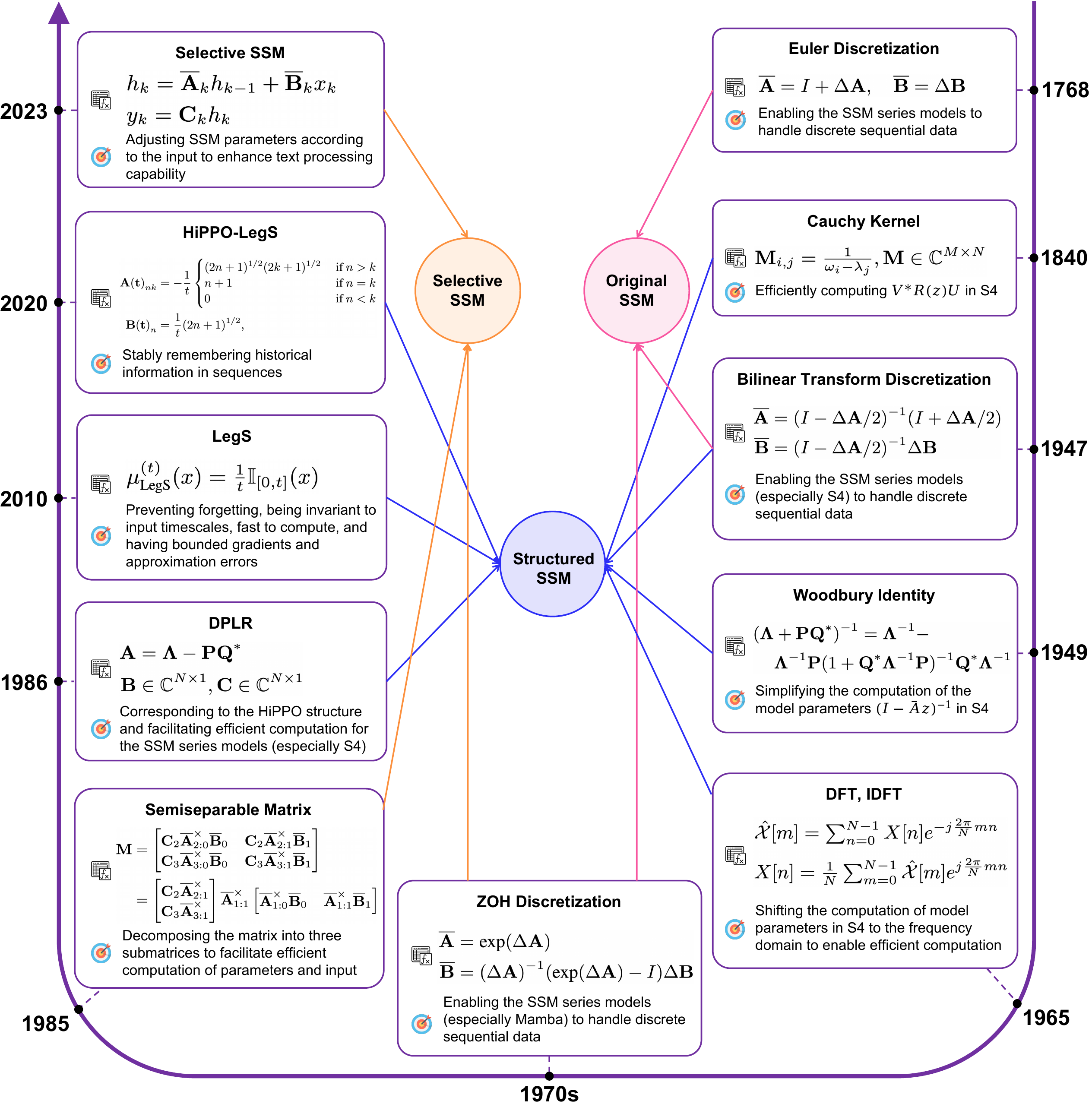}
\caption{Technologies of SSM Series and their corresponding relationships. Each technical description card includes the technique’s name, mathematical formulation, and objective.}
\label{fig:technology}
\end{figure}

Several key techniques have provided technical support and ensured the development of the SSM series. Methods such as Euler’s method~\citep{euler1794institutiones}, zero-order hold (ZOH), and bilinear transform~\citep{tustin1947method} discretization enable the transformation of SSMs from continuous\ybsun{-time} to discrete\ybsun{-time}, allowing these models to effectively handle discrete sequential data. To enhance the performance of SSMs, researchers \ybsun{ have introduced mathematical }techniques like LegS~\citep{hosny2010refined}, \ybsun{HiPPO}~\citep{gu2020hippo}, and selective SSM~\citep{dao2024transformers}. \ybsun{These techniques} enable stable retention of historical sequence information and significantly improve the models’ text processing capabilities. \ybsun{In terms of} efficiency, techniques such as DPLR~\citep{antoulas1986scalar}, DFT, IDFT~\citep{cooley1969finite}, semi-separable matrices~\citep{gohberg1985linear} lay the foundation for faster computations, and
tools like the Woodbury identity~\citep{woodbury1949stability} and Cauchy kernel~\citep{cauchy1840exercices} streamline the computational processes.
\ybsun{Together, these recent advancements greatly improve the computational efficiency of SSMs with little to no loss in performance.} Figure~\ref{fig:technology} presents these techniques and their corresponding relationships with different models in chronological order.

In this survey, 
\ybsun{We first introduce the mathematical formulation of SSMs in Section \ref{State Space Models: From Continuous To Discrete}. 
Next, Section \ref{S4: More Effective and Efficient} explores additional structures in SSMs with works such as S4, Section \ref{Mamba: More Practical and Powerful} introduces selectivity for SSMs leading with Mamba.}
These sections focus on detailing core techniques, summarizing optimization strategies in model architectures, and discussing theoretical analyses. Section~\ref{SSM and Other Model Architecture} explores the relationship between SSMs and other model structures, particularly Transformer, and examines the trends in SSM architecture optimization. Section~\ref{Application} highlights the applications of SSMs across various domains, including video processing, molecular modeling, speech and audio analysis, and so on.

\input{figures/mindmap}

%% file: figures/mindmap.tex
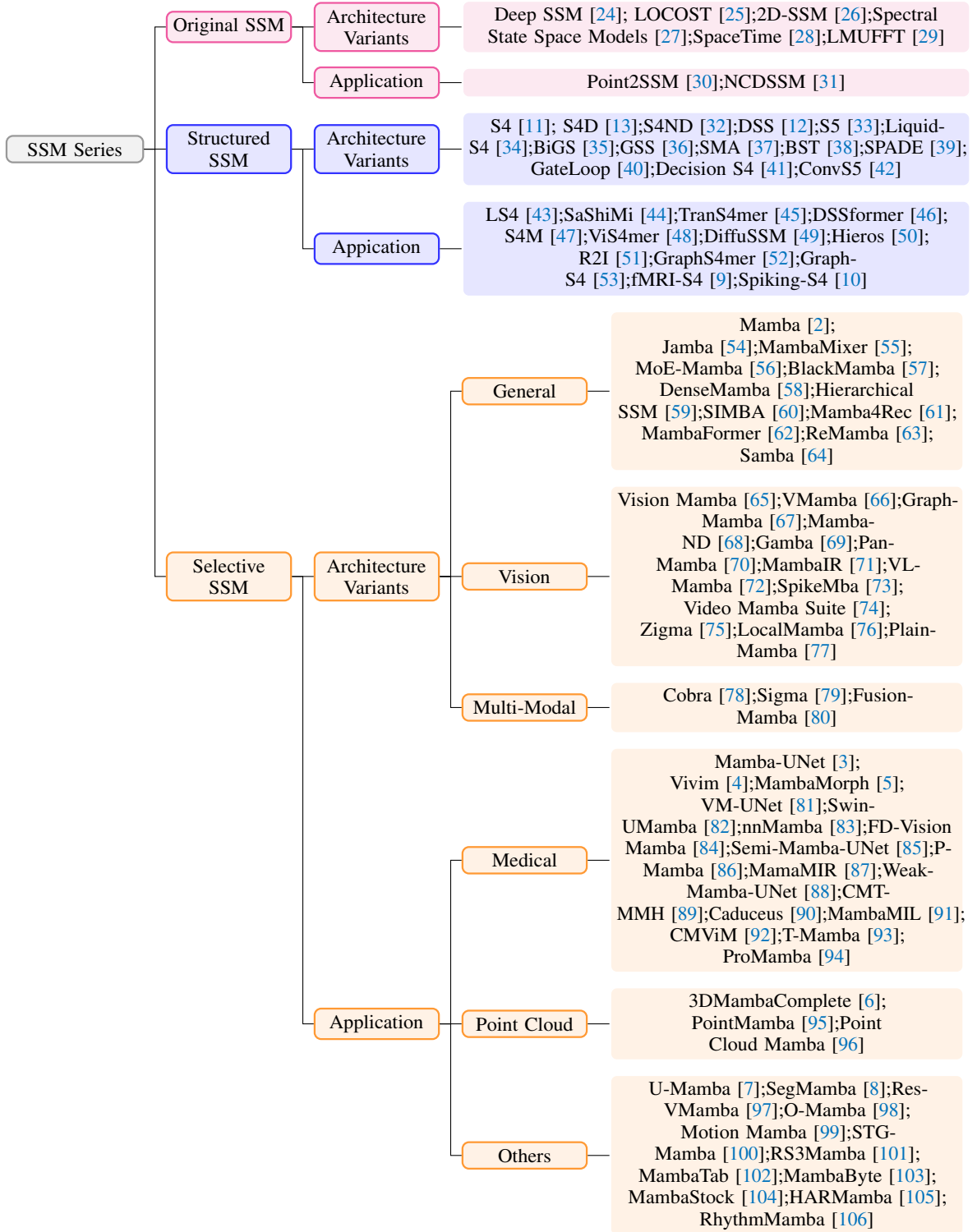
\begin{figure*}
\footnotesize
        \begin{forest}
            for tree={
                forked edges,
                grow'=0,
                draw,
                rounded corners,
                node options={align=center,},
                text width=2.7cm,
                s sep=6pt,
                calign=child edge, calign child=(n_children()+1)/2,
            },
            [SSM Series, fill=gray!45, parent
                [Original SSM, for tree={fill=green!45, o_ssm}
                    [Architecture Variants,  o_ssm
                        [Deep SSM~\citep{NEURIPS2018_5cf68969}; LOCOST~\citep{bronnec2024locost};2D-SSM~\citep{baron2024a};Spectral State Space Models~\citep{agarwal2024spectral};SpaceTime~\citep{zhang2023effectively};LMUFFT~\citep{10289082},
                        o_ssm_work]
                    ]
                    [Application, o_ssm
                        [Point2SSM~\citep{adams2024pointssm};NCDSSM~\citep{ansari2023neuralcontinuousdiscretestatespace},o_ssm_work]
                    ]
                ]
                [Structured SSM, for tree={fill=red!45,s4}
                    [Architecture Variants,  s4
                        [S4~\citep{gu2021efficiently}; S4D~\citep{gu2022parameterizationinitializationdiagonalstate};S4ND~\citep{nguyen2022s4ndmodelingimagesvideos};DSS~\citep{gupta2022diagonalstatespaceseffective};S5~\citep{smith2023simplifiedstatespacelayers};Liquid-S4~\citep{hasani2022liquid};BiGS~\citep{wang2023pretraining};GSS~\citep{mehta2022long};SMA~\citep{ren2023sparse};BST~\citep{fathi2023block-state};SPADE~\citep{zuo2022efficient};\\GateLoop~\citep{katsch2024gateloop};Decision S4~\citep{bardavid2023decisions4efficientsequencebased};ConvS5~\citep{smith2023convolutionalstatespacemodels}, s4_work]
                    ]
                    [Appication,  s4
                        [LS4~\citep{zhou2023deeplatentstatespace};SaShiMi~\citep{goel2022itsrawaudiogeneration};TranS4mer~\citep{islam2023efficientmoviescenedetection};DSSformer~\citep{saon2023diagonalstatespaceaugmented};\\S4M~\citep{chen2023neuralstatespacemodelapproach};ViS4mer~\citep{islam2023longmovieclipclassification};DiffuSSM~\citep{yan2023diffusionmodelsattention};Hieros~\citep{mattes2024hieroshierarchicalimaginationstructured};\\R2I~\citep{samsami2024masteringmemorytasksworld};GraphS4mer~\citep{tang2023modelingmultivariatebiosignalsgraph};Graph-S4~\citep{gazzar2022improvingdiagnosispsychiatricdisorders};fMRI-S4~\citep{elgazzar2022fmris4learningshortlongrange};Spiking-S4~\citep{du2024spikingstructuredstatespace}, s4_work]
                    ]
                ]
                [Selective SSM, for tree={fill=blue!45, mamba}
                    [Architecture Variants, mamba
                        [General[Mamba~\citep{gu2023mamba}; Jamba~\citep{lieber2024jamba};MambaMixer~\citep{behrouz2024mambamixer};\\MoE-Mamba~\citep{Pioro2024MoEMambaES};BlackMamba~\citep{Anthony2024BlackMambaMO};\\DenseMamba~\citep{He2024DenseMambaSS};Hierarchical SSM~\citep{Bhirangi2024HierarchicalSS};SIMBA~\citep{patro2024simbasimplifiedmambabasedarchitecture};Mamba4Rec~\citep{liu2024mamba4recefficientsequentialrecommendation};\\MambaFormer~\citep{park2024mambalearnlearncomparative};ReMamba~\citep{yuan2024remambaequipmambaeffective};\\Samba~\citep{ren2024sambasimplehybridstate},mamba_work]
                        ]
                        [Vision
                            [Vision Mamba~\citep{zhu2024visionmambaefficientvisual};VMamba~\citep{liu2024vmambavisualstatespace};Graph-Mamba~\citep{wang2024graphmambalongrangegraphsequence};Mamba-ND~\citep{li2024mambandselectivestatespace};Gamba~\citep{shen2024gambamarrygaussiansplatting};Pan-Mamba~\citep{he2024panmambaeffectivepansharpeningstate};MambaIR~\citep{guo2024mambairsimplebaselineimage};VL-Mamba~\citep{qiao2024vlmambaexploringstatespace};SpikeMba~\citep{li2024spikembamultimodalspikingsaliency};\\Video Mamba Suite~\citep{chen2024videomambasuitestate};\\Zigma~\citep{hu2024zigmaditstylezigzagmamba};LocalMamba~\citep{huang2024localmambavisualstatespace};Plain-Mamba~\citep{yang2024plainmambaimprovingnonhierarchicalmamba},mamba_work]
                        ]
                        [Multi-Modal
                        [Cobra~\citep{zhao2024cobraextendingmambamultimodal};Sigma~\citep{wan2024sigmasiamesemambanetwork};Fusion-Mamba~\citep{dong2024fusionmambacrossmodalityobjectdetection},mamba_work]
                        ]
                    ]
                    [Application, mamba
                        [Medical
                        [
                        Mamba-UNet~\citep{wang2024mambaunetunetlikepurevisual};\\Vivim~\citep{yang2024vivimvideovisionmamba};MambaMorph~\citep{guo2024mambamorphmambabasedframeworkmedical};\\VM-UNet~\citep{ruan2024vmunetvisionmambaunet};Swin-UMamba~\citep{liu2024swinumambamambabasedunetimagenetbased};nnMamba~\citep{gong2024nnmamba3dbiomedicalimage};FD-Vision Mamba~\citep{zheng2024fdvisionmambaendoscopicexposure};Semi-Mamba-UNet~\citep{ma2024semimambaunetpixellevelcontrastivepixellevel};P-Mamba~\citep{ye2024pmambamarryingperonamalik};MamaMIR~\citep{huang2024mambamirarbitrarymaskedmambajoint};Weak-Mamba-UNet~\citep{wang2024weakmambaunetvisualmambamakes};CMT-MMH~\citep{zhang2024motionguideddualcameratrackerendoscope};Caduceus~\citep{schiff2024caduceusbidirectionalequivariantlongrange};MambaMIL~\citep{yang2024mambamilenhancinglongsequence};\\CMViM~\citep{yang2024cmvimcontrastivemaskedvim};T-Mamba~\citep{hao2024tmambaunifiedframeworklongrange};\\ProMamba~\citep{xie2024promambapromptmambapolypsegmentation}, mamba_sub_work
                        ]]
                        [Point Cloud
                        [3DMambaComplete~\citep{li20243dmambacompleteexploringstructuredstate};\\PointMamba~\citep{liang2024pointmambasimplestatespace};Point Cloud Mamba~\citep{zhang2024pointcloudmambapoint},mamba_sub_work]
                        ]
                        [Others
                        [U-Mamba~\citep{ma2024umambaenhancinglongrangedependency};SegMamba~\citep{xing2024segmambalongrangesequentialmodeling};Res-VMamba~\citep{chen2024resvmambafinegrainedfoodcategory};O-Mamba~\citep{dong2024omambaoshapestatespacemodel};\\Motion Mamba~\citep{zhang2024motionmambaefficientlong};STG-Mamba~\citep{li2024stgmambaspatialtemporalgraphlearning};RS3Mamba~\citep{ma2024rs3mambavisualstatespace};\\MambaTab~\citep{ahamed2024mambatab};MambaByte~\citep{Wang2024MambaByteTS};\\MambaStock~\citep{shi2024mambastockselectivestatespace};HARMamba~\citep{li2024harmambaefficientlightweightwearable};\\RhythmMamba~\citep{zou2024rhythmmambafastremotephysiological},mamba_sub_work]
                        ]
                    ]
                ] 
            ]
        \end{forest}
            \caption{Typology of SSM Series. The “Architecture Variants” presents the common SSM architectures, while the “Application” shows the practical implementations of SSMs in real-world scenarios.}
            \label{fig:typo-prompt}
\end{figure*}

%% file: sections/2_ssm.tex



The concept of state space, along with the state space model, holds a significant historical backdrop.
The continuous-time state space model is a pivotal and versatile \ybsun{tool for describing dynamical systems. Early applications of state space models include} theoretical physics, communication signal processing, system control, and allied fields~\citep{findlay1904phase, bubnicki2005modern}.
Notably, Kalman~\citep{kalman1960new} employs state space models to characterize linear dynamic systems and resolve the optimal estimation problem.
By directly discretizing the continuous-time SSM, we obtain the discrete-time SSM, also referred to as the original SSM in this survey. 
The discrete-time SSM is capable of handling a series of discrete input data and has been increasingly gaining attention as a sequence model.

In this chapter, Section~\ref{Continuous-Time State Space Model} introduces the continuous-time State Space Model. Section~\ref{Discretization of Continuous-Time SSM} presents the discretization methods and the resulting discrete-time SSM. Section~\ref{Fundamental Properties of SSM} discusses several fundamental properties of SSMs, while Section~\ref{Models with Original SSMs} explores some model structures with the original SSM.

\subsection{Continuous-Time State Space Models}
\label{Continuous-Time State Space Model}

The classical linear continuous-time state space model (SSM) was first widely employed in control system theory~\citep{kalman1960new}. It describes dynamical systems through specific differential equations, and can be expressed as the following:
\begin{align}
    &h'(t) = \mathbf{A}(t)h(t) + \mathbf{B}(t)x(t), 
    \label{equ: continuous ssm1}\\
    &y(t) = \mathbf{C}(t)h(t) + \mathbf{D}(t)x(t), 
    \label{equ: continuous ssm2}
\end{align}
where $x(t) \in \mathbb{R}^{N_{\text{in}}}$ is the input vector signal, $h(t) \in \mathbb{R}^{N_{\text{s}}}$ is the hidden state vector signal, and $y(t) \in \mathbb{R}^{N_{\text{out}}}$ is the output vector signal. The matrices $\mathbf{A}(t)$, $\mathbf{B}(t)$, $\mathbf{C}(t)$, and $\mathbf{D}(t)$ represent the parameter matrices of the state space model.
Specifically, $\mathbf{A}(t) \in \mathbb{R}^{N_{\text{s}} \times N_{\text{s}}}$ is the system matrix, which describes how the current state influences its rate of change. $\mathbf{B}(t) \in \mathbb{R}^{N_{\text{s}} \times N_{\text{in}}}$, the control matrix, represents the effect of the input on the state change. $\mathbf{C}(t) \in \mathbb{R}^{N_{\text{out}} \times N_{\text{s}}}$, the output matrix, captures how the system states affect the output. $\mathbf{D}(t) \in \mathbb{R}^{N_{\text{out}} \times N_{\text{in}}}$, the feed-forward matrix, illustrates direct input-output relationships, circumventing system states.
Generally, in control systems or real-world physical systems, these parameters are either determined by the inherent characteristics of the system or acquired through conventional statistical inference methods.

\begin{figure}
    \centering
    \includegraphics[width=0.5\linewidth]{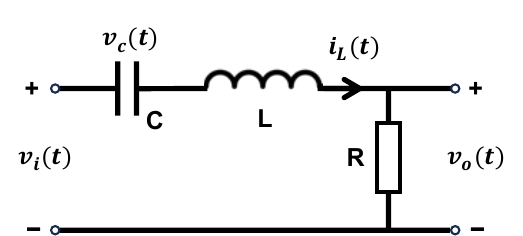}
    \caption{The illustration of a RC oscillator circuit, where $L$, $C$, and $R$ denote the inductance, capacitance and resistance, respectively. 
    }
    \label{fig:rc-oscillator}
\end{figure}

The classical continuous-time SSM can capture many systems in a wide range of subjects. For instance, it can model the RC oscillator circuit system shown in Figure~\ref{fig:rc-oscillator}, where $v$, $i$, $L$, $C$, and $R$ represent the voltage, current, inductance, capacitance and resistance, respectively.
Let the input signal be the system’s input voltage $[v_i(t)]=x(t)$, the output signal be the system’s output voltage $[v_o(t)]=y(t)$, and the two components of the system’s hidden state be the current in the inductor and the voltage across the capacitor $[i_L(t), v_c(t)]^T = h(t)$. 
Derived from fundamental physical principles such as Ohm's law and Kirchhoff's circuit laws, the state space model for this system can be expressed as:
\begin{align}
    &h'(t) = \mathbf{A}(t)h(t) + \mathbf{B}(t)x(t) = \bigg[\begin{matrix}
        -R & 0 \\
        0 & -1
    \end{matrix}\bigg]h(t)+[1]x(t), 
    \\
    &y(t) = \mathbf{C}(t)h(t) + \mathbf{D}(t)x(t) =  \bigg[\begin{matrix}
        -L\nabla_t & 0 \\
        0 & -1
    \end{matrix}\bigg]h(t)+[1]x(t).
\end{align}

SSM effectively describes how systems evolve over time. Much like tracking a moving object, the model uses its current position (state) to predict where it will go next. 
Since every change in the system follows clear mathematical rules, it is possible to predict future behavior or estimate hidden information from partial observations.
Furthermore, SSM is particularly suitable for describing complex systems, as it can simultaneously handle multiple inputs and outputs. 
For example, when tracking a rocket, inputs might include factors like fuel and wind speed, states could represent position and velocity, and outputs could correspond to the measurements recorded by the sensors.


\subsection{Discretization of Continuous-Time SSMs}
\label{Discretization of Continuous-Time SSM}
The above SSM described by Eq.~\ref{equ: continuous ssm1},\ref{equ: continuous ssm2} is termed the continuous-time SSM because $x(t)$, $h(t)$, and $y(t)$ represent continuous-time signals.
However, in most practical applications, the model inputs, hidden states, and outputs are instead discrete sequences, denoted as $x = (x_0, x_1, x_2, \dots)$, $h = (h_0, h_1, h_2, \dots)$, $y = (y_0, y_1, y_2, \dots)$. For instance, when processing natural language, the input and output of the model are token embeddings, which are discrete values rather than continuous functions.
Directly applying the continuous-time SSM to discrete data requires fitting the discrete data points into continuous signals, a complex and rarely used process. Instead, a more effective and practical approach is to discretize the SSM.

Essentially, the discrete-time SSM parameters $\overline{\mathbf{A}}$ and $\overline{\mathbf{B}}$ are expressed in terms of the continuous-time SSM parameters $\mathbf{A}$, $\mathbf{B}$, and an additional \ybsun{time-step} parameter $\Delta$.
And the discretization of SSMs generally involves addressing the ordinary differential equation (ODE) within the model representation.
The discrete-time SSM can be represented by the following difference equations:
\begin{align}
    &h_k = \overline{\mathbf{A}}h_{k-1} + \overline{\mathbf{B}}x_k, 
    \label{equ: discrete ssm1}\\
    &y_k = \mathbf{C}h_k + \mathbf{D}x_k,
    \label{equ: discrete ssm2}
\end{align}
where $x_k \in \mathbb{R}^{N_{\text{in}}}$, $h_k \in \mathbb{R}^{N_{\text{s}}}$ and $y_k \in \mathbb{R}^{N_{\text{out}}}$ denote one single data point of the input, hidden state, and output data sequences respectively. $\overline{\mathbf{A}} \in \mathbb{R}^{N_{\text{s}} \times N_{\text{s}}}$, $\overline{\mathbf{B}} \in \mathbb{R}^{N_{\text{s}} \times N_{\text{in}}}$, $\mathbf{C} \in \mathbb{R}^{N_{\text{out}} \times N_{\text{s}}}$, and $\mathbf{D} \in \mathbb{R}^{N_{\text{out}} \times N_{\text{in}}}$ are model parameters of the discrete-time SSM.

Notably, the parameter matrix $\mathbf{D}$, when researched as a sequence model, is often disregarded. This is because $\mathbf{D}$ is generally viewed as a skip connection and does not directly influence the state $h$. Usually, $\mathbf{D}$ is easy to compute, and can be approximated by other structures. Consequently,
Eq.~\ref{equ: discrete ssm1},\ref{equ: discrete ssm2} are commonly expressed as 
\begin{align}
    &h_k = \overline{\mathbf{A}}h_{k-1} + \overline{\mathbf{B}}x_k, 
    \label{equ: discrete ssm4}\\
    &y_k = \mathbf{C}h_k
    \label{equ: discrete ssm3}
\end{align}

Common discretization methods include Euler’s method~\citep{euler1794institutiones}, zero-order hold (ZOH), and the bilinear transform~\citep{gu2023modeling}:
\begin{align}
    &\textbf{(Euler)} \quad\quad \overline{\mathbf{A}} = I + \Delta \mathbf{A}, \quad\quad \overline{\mathbf{B}} = \Delta \mathbf{B}, \\
    &\textbf{(ZOH)} \quad\quad \overline{\mathbf{A}} = \text{exp}(\Delta \mathbf{A}), \quad\quad \overline{\mathbf{B}} = (\Delta \mathbf{A})^{-1}(\text{exp}(\Delta \mathbf{A}) - I)\Delta \mathbf{B}, \\
    &\textbf{(Bilinear)} \quad \overline{\mathbf{A}} = (I - \Delta \mathbf{A}/2)^{-1}(I + \Delta \mathbf{A}/2), \quad\quad \overline{\mathbf{B}} = (I - \Delta \mathbf{A}/2)^{-1}\Delta \mathbf{B}. \label{equ: Bilinear}
\end{align}

The discrete-time SSM can be used as a sequence model to model and process data that can be serialized.
For example, in natural language processing, given an input sequence $x = (x_0, x_1, x_2, \dots)$ where each $x_i$ represents the embedding vector of a linguistic token, the SSM can iteratively apply Eq.~\ref{equ: discrete ssm4},\ref{equ: discrete ssm3} to compute the hidden state-output pairs $(h_0, y_0), (h_1, y_1), \dots$ This recurrent computation process ultimately generates the output token representations $y = (y_0, y_1, y_2, \dots)$.
When both the input token embeddings and the corresponding desired output token representations are provided, the SSM can be trained to model the relationship between them. This establishes its capability for modeling sequential linguistic information.
In this paper, we focus on SSMs as sequence models
and use "SSM" to refer to the discrete-time SSM unless otherwise specified.

\subsection{Fundamental Properties of SSMs}
\label{Fundamental Properties of SSM}

The discrete-time SSM described by Eq.~\ref{equ: discrete ssm4},\ref{equ: discrete ssm3} can handle discrete sequential data and serve as a sequence model. We next discuss some fundamental properties of SSMs in this context.

A crucial property of the discrete-time SSM is its linearity.
Specifically, when emphasizing the input and output and simplifying Eq.~\ref{equ: discrete ssm4},\ref{equ: discrete ssm3} to $y = \text{SSM}(\overline{\mathbf{A}},\overline{\mathbf{B}},\mathbf{C})(x)$, it follows that 
\begin{equation}
\text{SSM}(\overline{\mathbf{A}},\overline{\mathbf{B}},\mathbf{C})(x^1 + x^2) = \text{SSM}(\overline{\mathbf{A}},\overline{\mathbf{B}},\mathbf{C})(x^1) + \text{SSM}(\overline{\mathbf{A}},\overline{\mathbf{B}},\mathbf{C})(x^2),
\end{equation}
where $x^1 = (x^1_0, x^1_1, x^1_2, \dots)$, $x^2 = (x^2_0, x^2_1, x^2_2, \dots)$ represent two different sets of input data. Another characteristic of linearity is that the model parameters $\overline{\mathbf{A}},\overline{\mathbf{B}},\mathbf{C}$ are independent of the input data $x$.
On one hand, this linearity enables optimizations for computational efficiency and parallelization, which will be discussed in Chapter~\ref{S4: More Effective and Efficient}. 
On the other hand, linear models often struggle with fitting complex data. Chapter~\ref{Mamba: More Practical and Powerful} will provide a detailed discussion of enabling input-dependent parameterization of model parameters $\overline{\mathbf{A}},\overline{\mathbf{B}},\mathbf{C}$, thereby transforming the SSM into a nonlinear architecture with enhanced selectivity.

Another fundamental property of SSM is that its expression can be derived into various forms. 
For instance, Eq.~\ref{equ: discrete ssm4},\ref{equ: discrete ssm3} naturally align with the representation format of recurrent models. These expressions can also lead to the derivation of convolutional forms, such that:
\begin{equation}
y = x * \overline{{\mathbf{K}} }   \quad where \quad x = (x_0, x_1, x_2, \dots),\overline{\mathbf{K}} = (\mathbf{C}\overline{\mathbf{B}}, \mathbf{C}\overline{\mathbf{AB}}, \mathbf{C}\overline{\mathbf{A}}^2\overline{\mathbf{B}}, \dots).
\label{equ: conv ssm}
\end{equation}
This convolutional structure creates the conditions necessary for the efficient computational algorithms introduced in Chapter~\ref{S4: More Effective and Efficient}.
This diversity of SSM expressions opens up flexibility in architectural modifications.

Furthermore, the SSM has several inherent limitations, including non-parallelizable computations, instability in capturing long sequence dependencies, and constrained approximation capacity.
These challenges can be addressed by incorporating structured parameters and selectivity, as will be elaborated in Chapters~\ref{S4: More Effective and Efficient} and~\ref{Mamba: More Practical and Powerful}. 
In this paper, we refer to the SSM described by Eq.~\ref{equ: discrete ssm4},\ref{equ: discrete ssm3} as the "original SSM" or the "vanilla SSM".

\subsection{Models with Original SSMs}
\label{Models with Original SSMs}

Despite its limitations, the original SSM remains widely used in some studies, including recent works, due to its simplicity and ease of implementation. This section presents some models that incorporate the original SSM. These model structures either embed the original discrete SSM within other architectures or integrate new computational modules into the SSM structure.
These modifications improve performance on tasks such as time series forecasting, long-text summarization, and two-dimensional data processing.

To accomplish long document abstractive summarization, Bronnec et al.~\citep{bronnec2024locost} construct an encoder layer with an SSM as its central component, termed the LOCOST. Starting from the convolutional representation of SSM, they introduce the Bidirectional SSM, which incorporates causal convolution and cross-correlation to simultaneously consider contextual information before and after each token in the sequence. This architecture builds upon the low computational complexity and superior long sequence handling of SSMs and further achieves improved context capture.
Baron et al.~\citep{baron20232} propose a 2-D SSM layer based on Roesser’s SSM model~\citep{givone1972multidimensional}. It expands states to capture both horizontal and vertical information, enhancing the SSM's understanding of two-dimensional data.
Agarwal et al.~\citep{agarwal2024spectral} apply the Spectral Filtering algorithm to learn linear dynamical systems. The proposed Spectral SSM employs fixed spectral filters that obviate the need for learning. It therefore effectively maps input sequences to a new representation space and captures long-term dependencies in sequential data. These features improve the model's prediction performance and stability, especially in sequences containing long-range dependencies.
Zhang et al.~\citep{zhang2023effectively} introduce a closed-loop prediction mechanism to SSMs, where the SSM at the decoder layer not only receives information from the encoder layer but also incorporates feedback from its own output. This design improves the model's flexibility and accuracy in handling long-term prediction tasks.

%% file: sections/3_s4.tex

\ybsun{The introduction of structured parameters to SSMs is considered to be one of the most remarkable recent advancements for sequence modeling.}
By \ybsun{the application of various structures to the parameters in SSMs}, the effectiveness and efficiency of \ybsun{these models} are improved. As one of the most prominent structured SSMs, S4 has garnered widespread attention and influence~\citep{gu2021efficiently}.
Within this chapter, we \ybsun{first discuss} the structures of model parameters introduced by S4 \ybsun{with an emphasis on its theoretical motivations. Next, we consider other structural mechanisms beyond S4. Lastly, we explore the role SSMs play as a part of more sophisticated models.}
Section~\ref{sec: overview} provides an overview of the structured SSM, especially S4, from an accessible perspective \ybsun{without delving into details.}
Sections~\ref{sec: effective hippo},~\ref{sec: efficient dplr} elaborate on S4 from a theoretical perspective, focusing on its effectiveness and efficiency, respectively.
Section~\ref{s4: other structures} summarizes \ybsun{alternative} structures to which the parameters of SSMs are restricted.
Section~\ref{s4: models embed s4 structure} reviews models that integrate structured SSM as a core component.

\subsection{Overview of Structured {State Space Models}}
\label{sec: overview}


\ybsun{The term structured SSM refers to SSMs where specific structures have been applied to their system dynamics. The exact structures can be formulated by a number of sophisticated methods such as initialization, parameterization or placing explicit constraints on parameters. }
For example, consider a model parameter matrix $\mathbf{W} \in \mathbb{R}^{N \times N}$. 
\ybsun{When its off-diagonal elements are restricted to zero, }$\mathbf{W}$ becomes a diagonal matrix and thus qualifies as a structured parameter.
This diagonal structure reduces storage overhead and computational cost.
In contrast to original \ybsun{unconstrained} SSMs,
structured SSMs such as S4 exhibit \ybsun{a number of desirable} mathematical properties, 
\ybsun{which greatly enhance their} effectiveness and efficiency.

Specifically, S4 \ybsun{demonstrated exceptional performance across a wide range of sequence modeling tasks. This is achieved by introducing a series of structural features: S4 first }leverages HiPPO~\citep{gu2020hippo} to constrain
parameters to the Legendre memory unit (discussed in detail in Section~\ref{sec: effective hippo}).
\ybsun{This specific structure } enables S4 to address challenges associated with \ybsun{memory retention for} long-range dependencies and alleviate issues such as gradient vanishing. Additionally, S4 employs the Diagonal Plus Low-Rank (DPLR)~\citep{antoulas1986scalar} structure (elaborated in Section~\ref{sec: efficient dplr}), which facilitates the use of efficient algorithms. \ybsun{These algorithms enable} parallel training and \ybsun{boost }computational efficiency \ybsun{and stability}. 
\ybsun{With the introduction of DPLR,} the computational complexity is reduced from $O(LN_{\text{s}}^2)$ to $O(L+N)$.
Beyond S4, alternative \ybsun{structured} SSM variants employ different structural constraints.
\ybsun{These modifications seek to refine and simplify the structured SSM framework and address the various inherent challenges in SSMs, }
including exponential memory decay, gradient vanishing, gradient explosion, instability in capturing long sequence dependencies, low training efficiency, and limited parallel computing capabilities.

\subsection{{Effectiveness: High-Order Polynomial Projection Operators (HiPPO)}}
\label{sec: effective hippo}

\ybsun{The traditional discrete-time SSM framework suffers from suboptimal performance, with} issues such as exponential memory decay and instability in capturing long sequence dependencies.
\ybsun{These issues }stem from the model's inability to effectively retain the input history.
\ybsun{S4 seeks to address these issues by }utilizing structured parameters, \ybsun{the specific structure of }which is derived through the HiPPO framework~\citep{gu2020hippo}.

Conceptually speaking, \ybsun{HiPPO addresses the following }core problem: How to efficiently reconstruct the history of input $x$ using state $h$? \ybsun{This problem can be formulated precisely as an} online function approximation problem, \ybsun{which is} investigated in HiPPO. In this subsection, we provide a detailed exposition of the HiPPO framework, discuss its specific instances and advantages, and explore the relationship between HiPPO and S4.

Gu et al.~\citep{gu2020hippo} specifically define memory through the online function approximation problem, which involves representing a function by storing coefficients of a set of basis functions. Mathematically, \ybsun{given} a function $f(t) \in \mathbb{R}$ on $t \geq 0$
and a \ybsun{probability} measure $\mu$, the distance between two functions $f$ and $g$ can be expressed as $\left \langle f,g \right \rangle_{\mu} =\int_{0}^{\infty } f(x)g(x)d\mu(x)$, and the norm of function $f$ can be defined as $\left \| f \right \|_{L_2(\mu)}  = \left \langle f,f \right \rangle_{\mu}^{1/2}$. \ybsun{Moreover,  the cumulative history is typically represented as $f_{\le t}:=f(s)|_{s\le t} $.} \
\ybsun{Solving the online approximation problem in this setup}
involves seeking $g^{(t)}\in\mathcal{G}$, typically a subspace spanned by orthogonal bases, that minimizes $ \| f_{\le t}-g^{(t)}  \|_{L_2(\mu^{(t)})}$. The HiPPO framework aims to optimize this approximation at every timestep t. 


Specifically, given an input function \ybsun{$f(t): \mathbb{R}^+ \rightarrow \mathbb{R}$} on $t \geq 0$  and a time-varying measure family $\mu^{(t)}$, the HiPPO framework consists of the following three steps: 
\begin{enumerate}
\item \textbf{Obtain the Set of Bases.} \ \ybsun{The first step in HiPPO is to }generate a suitable \ybsun{set of} basis functions $ \{ g_n^{(t)}  \}_{n\in \{ 0,1,\dots,N-1  \} } $ according to $\mu^{(t)}$.
$ \{ g_n^{(t)}  \} $ should satisfy $ \langle g_n^{(t)}, g_m^{(t)} \rangle_{\mu^{(t)}} = \lambda _n^2\delta _{n,m}$, where $\delta$ is the impulse function.
$\lambda _n$ is typically set to $\pm 1$ to ensure $ \{ g_n^{(t)}  \}$ forms the orthogonal bases. This set of bases spans an N-dimensional subspace $\mathcal{G}$.
\item \textbf{Calculate the Optimal Coefficients.} \ \ybsun{Given the generated basis set $ \{ g_n^{(t)} \}$, the next step is finding the optimal representation of input signal $f_{\le t}:=f(s)|_{s\le t} $ in the form of a polynomial $g^{(t)}= {\sum_{n=0}^{N-1}}c_n(t)g_n^{(t)}$. Here, $c_n(t)$ denotes the optimal coefficients to construct polynomial $g^{(t)}=\text{argmin}_{g\in\mathcal{G}} \| f_{\le t}-g  \|_{\mu^{(t)}}$.  
With the orthogonality of the bases, the optimal coefficients can be computed through the following,
\begin{equation}
c_n(t)=\left \langle f_{\le t},g_n^{(t)} \right \rangle_{\mu^{(t)}}.
\label{equ: c projection}
\end{equation}
The HiPPO paper denotes the results of the calculations, $g^{(t)}$ and $c(t) := (c_n(t))_{0\le n< N }$ as $\text{proj}_t$ and $\text{coef}_t$, respectively.
}


\item \textbf{Differentiate Equation~\ref{equ: c projection} and Yield the ODE.} \ 
\ybsun{The last equation shows how to calculate $c(t)$ given a signal $f(t)$. However, in a continuous-time state space system, $c(t)$ denotes the state and $f(t)$ denotes the input which changes over time. In order to establish a closer connection with the expression of SSM (Equation~\ref{equ: continuous ssm1}), it is necessary to find the optimal matrices $\mathbf{A}$ and $\mathbf{B}$, Furthermore, $\mathbf{A}$ and $\mathbf{B}$ should remain independent of $f(t)$. This can be achieved by taking the derivative of the previous equation.
Given the Equation~\ref{equ: c projection}, the differentiation of state $c(t)$ is calculated with respect to the input function $f(t)$. This is depicted by the following ordinary differential equation (ODE):}

\begin{equation}
\frac{d}{dt}c(t) = \mathbf{A}(t)c(t) + \mathbf{B}(t)f(t),
\label{equ: c ode}
\end{equation}
where $\mathbf{A}(t) \in \mathbb{R}^{N \times N}$ and $\mathbf{B}(t) \in \mathbb{R}^{N \times 1}$ often remain independent of $f(t)$ and are solely related to the measure $\mu^{(t)}$ and the orthogonal subspace $\mathcal{G}$. Equation~\ref{equ: c ode} offers a general strategy for computing coefficients $c(t)$.
\end{enumerate}
In essence, HiPPO can be conceptualized as a system: given a function $f(t)$ and a measure $\mu^{(t)}$, it outputs the optimal representation coefficients $c(t)$ of $f(t)$ under $\mu^{(t)}$.
It is worth noting that, in most cases, researchers are primarily interested in the dynamic relationships \ybsun{between $f(t)$ and }$c(t)$, which are represented by $\mathbf{A}(t)$ and $\mathbf{B}(t)$. Hence, HiPPO can also be viewed as a black box: given a measure $\mu^{(t)}$, it outputs the corresponding structured parameter matrices $\mathbf{A}(t)$ and $\mathbf{B}(t)$ satisfying the OED Equation~\ref{equ: c ode} and yielding the optimal representation coefficients $c(t)$. 


\ybsun{
Given specific measure, the HiPPO framework is able to generate optimal state-space equations. We discuss two notable SSM-related HiPPO instances, LegT and LegS.
}
The first \ybsun{variant} entails the use of the translated Legendre (LegT) measures.
LegT can be expressed as $\mu^{(t)}_\text{LegT}(x)=\frac{1}{\theta}\mathbb{I}_{\left [ t-\theta,t \right]}(x) $, where $\theta$ denotes the length of the history being memorized.
This measure assigns uniform weights to nearby histories.
Through the HiPPO framework, the structured parameter matrices $\mathbf{A}$ and $\mathbf{B}$ corresponding to LegT can be \ybsun{determined as}
\begin{equation}
\mathbf{A}_{nk}=-\frac{1}{\theta}\begin{cases}
 (-1)^{n-k}(2n+1) & \text{ if } n\ge k \\
 2n+1 & \text{ if } n\le k
\end{cases}, 
\quad \mathbf{B}_{n}=\frac{1}{\theta}(2n+1)(-1)^n,
\end{equation}
which is exactly the parameter matrices of the Legendre Memory Unit (LMU)~\citep{voelker2019legendre}, an improved model of RNN.

The second example comprises the scaled Legendre (LegS) measures~\citep{hosny2010refined}, denoted as $\mu^{(t)}_\text{LegS}(x)=\frac{1}{t}\mathbb{I}_{\left [ 0,t \right]}(x)$, for which the corresponding structured parameter matrices $\mathbf{A}$ and $\mathbf{B}$ satisfy: 
\begin{equation}
t\mathbf{A}(t)_{nk}=-\begin{cases}
 (2n+1)^{1/2}(2k+1)^{1/2} & \text{ if } n > k \\
 n+1 & \text{ if } n = k \\
 0 & \text{ if } n < k
\end{cases}, 
\quad t\mathbf{B}(t)_{n}=(2n+1)^{1/2},
\label{equ: legs a b}
\end{equation}
obtained through the HiPPO framework.
LegS is an optimization over LegT. It scales the window over time and assigns uniform weights to all history. \ybsun{Advantages of LegS include} mitigation of forgetting, invariance to input timescale, faster computations, and bounded gradients and approximation errors.

The HiPPO framework outputs the structured parameter matrices $\mathbf{A}(t)$ and $\mathbf{B}(t)$ that satisfy Equation~\ref{equ: c ode} and yield the optimal coefficients $c(t)$ for representing $f(t)$. In other words, setting the same $\mathbf{A}(t)$ and $\mathbf{B}(t)$ or similar parameter matrices in Equation~\ref{equ: continuous ssm1} enables the state $h(t)$ to optimally represent the input $x(t)$, thereby achieving \ybsun{the memorization} of history. 

\ybsun{S4 utilizes the specific structures generated by HiPPO to help with its long-range dependencies. Although the matrices in Equation~\ref{equ: legs a b} are optimal in theory, empirical results have shown that making parameters $\mathbf{A}(t)$ and $\mathbf{B}(t)$ trainable can further improves performance in general.}
In S4, if the parameters $\mathbf{A}$ and $\mathbf{B}$ are not trained, they are restricted to specific structures through HiPPO. However, training the parameters generally yields better results, hence HiPPO is commonly used for initialization. Typically, initialization often involves using the $t\mathbf{A}(t)$ and $t\mathbf{B}(t)$ corresponding to the LegS (i.e. Equation~\ref{equ: legs a b}) and discretizing through Equation~\ref{equ: Bilinear} in S4.

\subsection{{Efficiency:  Diagonal Plus Low-Rank}}
\label{sec: efficient dplr}
Another crucial challenge addressed by S4 is \ybsun{computational efficiency}. 
\ybsun{The complexity of the calculating the original SSM in Eq.~\ref{equ: discrete ssm4},\ref{equ: discrete ssm3} is $O(LN^2)$. And its training process lacks parallelizability.} 
S4 \ybsun{overcomes this limitation by introducing} of a special structure named Diagonal Plus Low-Rank (DPLR)~\citep{antoulas1986scalar} and \ybsun{employing a series of algorithms designed }for parallel training and efficient computation. 

The SSM with DPLR structure satisfies $\mathbf{A} = \mathbf{\Lambda} - \mathbf{P}\mathbf{Q}^*, \mathbf{B}\in \mathbb{C}^{N\times1},\mathbf{C}\in \mathbb{C}^{1\times N}$, where $\mathbf{\Lambda} \in \mathbb{C}^{N\times N}$ is diagonal \ybsun{and} $\mathbf{P},\mathbf{Q}\in \mathbb{C}^{N\times1}$ \ybsun{is low-rank}.
The corresponding algorithm begins with the convolutional representation of SSM: by recursively applying Equation~\ref{equ: discrete ssm1},\ref{equ: discrete ssm3}, $y_k = \mathbf{C}\overline{\mathbf{A}}^k\overline{\mathbf{B}}x_0 + \mathbf{C}\overline{\mathbf{A}}^{k-1}\overline{\mathbf{B}}x_1 + \dots + \mathbf{C}\overline{\mathbf{AB}}x_{k-1} + \mathbf{C}\overline{\mathbf{B}}x_{k}$ can be derived
This equation leads to 
\begin{equation}
y = \overline{\mathbf{K}}*x, \quad \overline{\mathbf{K}} \in \mathbb{R}^L := (\mathbf{C}\overline{\mathbf{A}}^n\overline{\mathbf{B}})_{n \in [L]}=(\mathbf{C}\overline{\mathbf{B}}, \mathbf{C}\overline{\mathbf{AB}}, \dots ,\mathbf{C}\overline{\mathbf{A}}^{L-1}\overline{\mathbf{B}})
\label{equ: conv with length L}
\end{equation}
where $x = (x_0, x_1, \dots, x_{L-1})$ is an input of length L. In Equation~\ref{equ: conv with length L}, the complexity of computing $\overline{\mathbf{K}}$ is $O(LN^2)$, making the efficient computation of $\overline{\mathbf{K}}$ a central aspect of the algorithm.

The first pivotal step of this algorithm involves shifting the computation of $\overline{\mathbf{K}}$ to the frequency domain by computing its Discrete Fourier Transform (DFT)~\citep{cooley1969finite}.
This can be denoted as
\begin{equation}
\hat{\mathcal{K}}_L(z;\overline{\mathbf{A}},\overline{\mathbf{B}},\mathbf{C}) \in \mathbb{C} := 
\sum_{n=0}^{L-1}\bar{\mathbf{C}}\overline{\mathbf{A}}^n\overline{\mathbf{B}}z^n
= \bar{\mathbf{C}}(\mathbf{I}-\overline{\mathbf{A}}^Lz^L)(\mathbf{I}-\overline{\mathbf{A}}z)^{-1}\overline{\mathbf{B}}
= \tilde{\mathbf{C}}(\mathbf{I}-\overline{\mathbf{A}}z)^{-1}\overline{\mathbf{B}},
\end{equation}
where $\bar{\mathbf{C}}$ denotes the conjugate of $\mathbf{C}$, $z \in \Omega = \left \{ \text{exp}(-2\pi i\frac{k}{L}):k\in[L] \right \}$, $\tilde{\mathbf{C}} = \bar{\mathbf{C}}(\mathbf{I}-\overline{\mathbf{A}}^Lz^L) = \bar{\mathbf{C}}(\mathbf{I}-\overline{\mathbf{A}}^L)$.
After obtaining the DFT, $\overline{\mathbf{K}}$ can be computed using the Fast Fourier Transform (FFT) algorithm, requiring $O(L\text{log}L)$ operations. It's worth noting that in S4, $\tilde{\mathbf{C}}$ is directly learned via parameterization, obviating the need for computing $\tilde{\mathbf{C}} = \bar{\mathbf{C}}(\mathbf{I}-\overline{\mathbf{A}}^L)$. 
\ybsun{This step replaces matrix power operations with efficient low-rank matrix inversions, presented as the subsequent step.}

The second integral step of this algorithm simplifies the computation of $(\mathbf{I}-\overline{\mathbf{A}}z)^{-1}$ through using the Woodbury identity~\citep{woodbury1949stability}. In the low-rank scenario, the Woodbury identity is
\begin{equation}
(\mathbf{\Lambda}+\mathbf{PQ}^*)^{-1} = \mathbf{\Lambda}^{-1} - \mathbf{\Lambda}^{-1}\mathbf{P}(1+\mathbf{Q}^*\mathbf{\Lambda}^{-1}\mathbf{P})^{-1}\mathbf{Q}^*\mathbf{\Lambda}^{-1},
\end{equation}
where $\mathbf{\Lambda} \in \mathbb{C}^{N\times N}$ is diagonal, $\mathbf{P},\mathbf{Q}\in \mathbb{C}^{N\times1}$, and $(\mathbf{\Lambda}+\mathbf{PQ}^*)^{-1}$ is invertible.
Applying Equation~\ref{equ: Bilinear},
\begin{equation}
\hat{\mathcal{K}}_L(z;\overline{\mathbf{A}},\overline{\mathbf{B}},\mathbf{C}) = 
\tilde{\mathbf{C}}(\mathbf{I}-\overline{\mathbf{A}}z)^{-1}\overline{\mathbf{B}} =
\frac{2}{1+z}\tilde{\mathbf{C}}\left (\frac{2}{\Delta}\frac{1-z}{1+z}-\mathbf{A}\right )^{-1}\mathbf{B}
\end{equation}
can be derived.
By leveraging the Woodbury identity and $\mathbf{A} = \mathbf{\Lambda} - \mathbf{P}\mathbf{Q}^*$ in the DPLR structure,
\begin{equation}
\begin{aligned}
    \hat{\mathcal{K}}_L(z;\overline{\mathbf{A}},\overline{\mathbf{B}},\mathbf{C}) &= \frac{2}{1+z}\tilde{\mathbf{C}}\left (\frac{2}{\Delta}\frac{1-z}{1+z}-\mathbf{\Lambda}+\mathbf{P}\mathbf{Q}^*\right )^{-1}\mathbf{B} \\
    &= \frac{2}{1+z}\left [ \tilde{\mathbf{C}}\mathbf{R}(z)\mathbf{B}-\tilde{\mathbf{C}}\mathbf{R}(z)\mathbf{P}(1+\mathbf{Q}^*\mathbf{R}(z)\mathbf{P})^{-1}\mathbf{Q}^*\mathbf{R}(z)\mathbf{B} \right ],
\end{aligned}
\end{equation}
where $\mathbf{R}(z) = \left ( \frac{2}{\Delta}\frac{1-z}{1+z}-\mathbf{\Lambda} \right )^{-1}$.
This step transforms the computation of $(\mathbf{I}-\overline{\mathbf{A}}z)^{-1}$ into the inversion of the diagonal matrix, streamlining the calculation process.

The final step of this algorithm focuses on efficiently computing $\mathbf{V}^*\mathbf{R}(z)\mathbf{U}$ using the Cauchy kernel~\citep{cauchy1840exercices}. Leveraging the previous steps, the computation of $\overline{\mathbf{K}}$ is transformed into calculating four expressions in the format of $\mathbf{V}^*\mathbf{R}(z)\mathbf{U}$ (i.e. $\tilde{\mathbf{C}}\mathbf{R}(z)\mathbf{B}$, $\tilde{\mathbf{C}}\mathbf{R}(z)\mathbf{P}$, $\mathbf{Q}^*\mathbf{R}(z)\mathbf{P}$, $\mathbf{Q}^*\mathbf{R}(z)\mathbf{B}$). Given $\mathbf{V}^*\mathbf{R}(z)\mathbf{U} =  {\textstyle \sum_{n}}\frac{v_n^*u_n}{z-\lambda_n}$, computing $\mathbf{V}^*\mathbf{R}(z)\mathbf{U}$ for all $z \in \Omega = \left \{ \text{exp}(-2\pi i\frac{k}{L}):k\in[L] \right \}$ entails a Cauchy matrix-vector multiplication.
This process has been extensively researched and requires only $\tilde{O}(L+N)$ operations.

S4 constrains the parameter matrices to adhere to the DPLR structure. 
\ybsun{However, the limited expressivity of DPLR does not compromise the effectiveness of models such as S4 since techniques such as the HiPPO framework inherently satisfy this structure.}
Specifically, the HiPPO matrices adhere to the Normal Plus Low-Rank (NPLR) structure: $\mathbf{A} = \mathbf{V}\mathbf{\Lambda}\mathbf{V}^* - \mathbf{P}\mathbf{Q}^* = \mathbf{V}(\mathbf{\Lambda} - (\mathbf{V}^*\mathbf{P})(\mathbf{V}^*\mathbf{Q})^*)\mathbf{V}^*$, where $\mathbf{V}\in \mathbb{C}^{N\times N}$ is unitary, $\mathbf{\Lambda}$ is diagonal, and $\mathbf{P},\mathbf{Q}\in \mathbb{R}^{N\times r}$. Particularly, the HiPPO matrices corresponding to LegS satisfy $r=1$. In this scenario, the NPLR structure is unitarily equivalent to DPLR.

In summary, S4 introduces the DPLR structure, shifts computations from time to frequency domain using the SSM's convolutional expression, and simplifies calculations through the Woodbury identity and Cauchy kernel. It achieves parallel computation and reduces training complexity to $O(N(\tilde{N}+\tilde{L}))$.


\input{sections/3_s4_sub1}


%% file: sections/3_s4_sub1.tex


\subsection{{Additional} Structural Constraints for Structured SSMs}
\label{s4: other structures}

Sections~\ref{sec: effective hippo} and~\ref{sec: efficient dplr} introduce two types of structured parameters in S4: the LegS structure derived from the HiPPO framework and the DPLR structure. In addition to these two types of structural constraints on model parameters, several other structural constraints have been proposed for structured SSMs. 
In this section, we examine additional structured parameters and their corresponding Structured SSMs, specifically S4D~\citep{gu2022parameterizationinitializationdiagonalstate} and DSS~\citep{gupta2022diagonalstatespaceseffective}. A comparative analysis of these two structured SSMs is also presented.

The Diagonal State Space (DSS)~\citep{gupta2022diagonalstatespaceseffective} removes the low-rank component of the DPLR structure, retaining only the diagonalizable term $\mathbf{\Lambda}$.
\ybsun{Compared to S4, DSS further reduces computational complexity with no significant loss of performance.}
Empirically, it achieves an average accuracy of 81.88 across six tasks in the Long Range Arena (LRA) benchmark~\citep{tay2020long}, closely rivaling the 80.21 of S4.
Its kernel from Eq.~\ref{equ: conv ssm} can be derived and denoted as:
\begin{align}
\mathbf{K} &= \bar{\mathbf{K}}_{\mathbf{\Delta}, L}\left(\mathbf{\Lambda},(1)_{1 \leqslant i \leqslant N}, \widetilde{w}\right) = \widetilde{w} \cdot \mathbf{\Lambda}^{-1}\left(e^{\mathbf{\Lambda} \mathbf{\Delta}}-I\right) \cdot \text{elementwise-exp}(\mathbf{P}),
\label{equ: dss}
\end{align}
where $\mathbf{K} \in \mathbb{R}^{1 \times L}$ represents the kernel of length $L$ with a sampling interval $\mathbf{\Delta}>0$. $\widetilde{w} \in \mathbb{C}^{1 \times N}$ is the parameter matrix. Matrix $\mathbf{P} \in \mathbb{C}^{N \times L}$ is defined as $\mathbf{P}_{i, k}=\lambda_i k \mathbf{\Delta}$. $\mathbf{\Lambda}$ is the diagonal matrix consisting of $\lambda_1, \ldots, \lambda_N$, with all $\lambda_i \neq 0$. The DSS eliminates the need for matrix powers and instead only requires a structured matrix-vector product.

S4D~\citep{gu2022parameterizationinitializationdiagonalstate} restricts the parameter matrix $\mathbf{A}$ to a diagonal structure, and its kernel computation method is simplified: 
\begin{align}
   \mathbf{K}=(\overline{\mathbf{B}}^{\top} \circ \mathbf{C}) \cdot \mathcal{V}_L(\overline{\mathbf{A}}) \quad \text{where} \quad \mathcal{V}_L(\overline{\mathbf{A}})_{n, \ell}=\overline{\mathbf{A}}_n^{\ell},
\end{align}
where $\mathbf{A}$ is a diagonal matrix, $\circ$ signifies the Hadamard product, and $\mathcal{V}$ denotes a Vandermonde matrix that shares the same $O(N+L)$ complexity with the Cauchy kernel used in the S4 algorithm.

\ybsun{Compared to S4, DSS and S4D also varies in terms of methods of }discretization. DSS adopts ZOH and S4D can use either ZOH or Bilinear discretization method.
During the training process, the real parts of the diagonal matrix's elements might turn positive. This change possibly leads to instability, especially when facing longer input sequences.
To circumvent this, DSS imposes constraints to ensure the real part of $\mathbf{\Lambda}_{Re}$ remains negative or substitutes the elementwise $\exp(\mathbf{P})$ (as in Eq.~\ref{equ: dss}) with $\operatorname{row}-\operatorname{softmax}(P)$.
The operation of $\operatorname{row}-\operatorname{softmax}(P)$ normalizes each row of elementwise $\exp(\mathbf{P})$ by the sum of its elements.
However, the softmax kernel adds complexity, faces challenges with varying input sequence lengths, and incurs additional memory costs. S4D also imposes constraints on the real component $\mathbf{A}_{Re}$ of its matrix $\mathbf{A}$, keeping it negative either by encapsulating it within an exponential function $\mathbf{A} = -\exp(\mathbf{A}_{Re}) + i \cdot \mathbf{A}_{Im}$ or by employing a negative-bounding activation function such as ReLU. Differing from S4D, which may either freeze the $\mathbf{B}$ matrix or train both $\mathbf{B}$ and $\mathbf{C}$ independently, DSS directly parameterizes and jointly trains the product of $\mathbf{B}$ and $\mathbf{C}$ as $\widetilde{w}$. In essence, S4D provides a comprehensive array of optimization strategies for diagonal state space-based models, catering to the varied demands of these algorithms. The following table presents an overview of the configuration differences between DSS and S4D:

\begin{center}
\begin{tabular}{  m{4cm}  m{4.5cm}  m{4.5cm}  }
\toprule
  & DSS & S4D \\
\midrule
Discretization Method & Zero-Order Hold (ZOH) & ZOH or Bilinear \\
\midrule
Training Constraints & Negative $\mathbf{\Lambda}_{Re}$ or softmax kernel & Negative $\mathbf{A}_{Re}$ (via exp or ReLU) \\
\midrule
Trainable $\mathbf{B}$, $\mathbf{C}$ matrices & Jointly trained $\mathbf{B} \circ \mathbf{C}$ as $\widetilde{w}$ & Independently trained $\mathbf{B}$, $\mathbf{C}$ or $\mathbf{C}$ alone \\
\bottomrule
\end{tabular}
\label{tab: dss and s4d}

\end{center}


\subsection{{Integration of Structured SSMs with Other Models}}
\label{s4: models embed s4 structure}

Combining the structured SSM and other models or methods has allowed researchers to exploit the long-context abilities and efficiency of structured SSM in diverse scenarios. This section introduces methods for integrating structured SSM with Liquid-Time Constant SSMs (LTCs), gating mechanisms, and transformer architectures.

Hasani et al.~\citep{hasani2022liquid} scale LTCs to long sequences by adopting reparameterization strategies used in S4 and showing that the additional liquid kernel can be efficiently computed based on the S4 kernel. This formulation combines the efficiency of S4s with the input-dependent state transitions of LTCs, achieving state-of-the-art results on several sequence modeling tasks.

To fuse S4-based models with gating mechanisms, Mehta et al.~\citep{mehta2022long} replace the attention module in a Gated Attention Unit (GAU) with a simplified DSS module. The resulting Gated State Space (GSS) layer performs on par with transformer-based models, is more efficient than DSS models, and showcases promising length generalization abilities. In Ren et al.~\citep{ren2023sparse}, an SSM module is one of the core components of Sparse Modular Activation (SMA) implementation in the proposed model SeqBoat. The SSM produces state representations that control the sparse activation of the subsequent GAU. Wang et al.~\citep{wang2023pretraining} use bidirectional pairs of S4D modules to replace attention in both transformers and gated units, achieving pretraining without attention. The latter configuration achieves performance similar to BERT. Katsch~\citep{katsch2024gateloop} introduces data-controlled input, output, and hidden state gates, arriving at a unified and generalized formulation of linear recurrent models such as S4. Compared with vanilla S4, the proposed GateLoop improves the model's expressiveness, leading to better autoregressive language modeling performance.

Methods incorporating S4 into transformers commonly use S4 to provide global context while lowering the quadratic computational complexity of attention mechanisms, thus producing architectures practical for modeling long sequences. Zuo et al.~\citep{zuo2022efficient} introduce the State Space Augmented Transformer (SPADE), which augments efficient transformers with a global context. SPADE's bottommost global layer integrates the outputs of both an S4 module and a local attention module. Fathi et al.~\citep{fathi2023block-state} propose a hybrid Block-State Transformer (BST) layer, which combines self-attention over input embeddings with cross-attention between inputs and context states. The context states, obtained from SSMs, allow the BST to maintain a grasp of the full global context despite only attending to short subsequences.

%% file: sections/4_mamba.tex
In addition to incorporating structured parameters, another key optimization strategy for the SSM series involves \ybsun{the introduction of} selectivity: allowing the model to dynamically prioritize specific segments of the input sequence. Mamba~\citep{gu2023mamba} is a representative example of a model centered on Selective SSM, and its introduction has garnered significant attention for the SSM series. Mamba demonstrates marked improvements in both model capability and task performance, driving its widespread adoption in real-world applications.
In this chapter, Section~\ref{sec: Overview of Mamba} provides an overview of selective SSM.
Section~\ref{sec: selective} elaborates on relevant selective mechanisms.
Sections~\ref{sec: Efficient: Hardware-aware State Expansion} and~\ref{sec: Efficient: Semiseparable Matrices} introduce optimizations in computational efficiency for the selective SSM in Mamba and Mamba2~\citep{dao2024transformers}, respectively. 
Section~\ref{sec: mamba Model Architecture Optimization} discusses the overall framework, core components, and variants of Mamba, while Section~\ref{sec: mamba Theoretical Analysis} presents analyses on selective SSM.


\subsection{Overview of Selective State Space Models}
\label{sec: Overview of Mamba}

Gu and Dao~\citep{gu2023mamba} introduced selectivity mechanisms into SSM to enhance its capability in processing input sequences. The proposed selective SSM, or S6, adjusts model parameters in response to input sequences, enabling dynamic prioritization of specific input segments. In the selective SSM, parameters $\overline{\mathbf{A}}$, $\overline{\mathbf{B}}$, and $\mathbf{C}$ become context-dependent variables that adjust based on input characteristics. We discuss this mechanism in detail in the next section.

The selective SSM, combined with additional components including projection layers, activation functions, and one-dimensional convolutional layers, forms the foundation of the influential SSM series, which includes Mamba. The framework of Mamba will be detailed in Section~\ref{sec: mamba Model Architecture Optimization}. As the computational core of Mamba and its derivatives, selective SSM enhances data modeling capabilities and introduces selective attention mechanisms that enable token prioritization.


\subsection{Effectiveness: Selectivity}
\label{sec: selective}

Mathematically, selectivity is implemented by dynamically deriving the parameters $\overline{\mathbf{A}}$, $\overline{\mathbf{B}}$, and $\mathbf{C}$ (denoted as $\overline{\mathbf{A}}_k$, $\overline{\mathbf{B}}_k$, and $\mathbf{C}_k$ in the selective SSM) from the input.
Specifically, for the data $x_k$ of the $k$-th input token, $x_k$ is processed through a linear network to generate parameters $\Delta_k$, $\mathbf{B}_k$, and $\mathbf{C}_k$. These are subsequently transformed using the zero-order hold (ZOH) discretization method, yielding $\overline{\mathbf{A}}_k = \text{exp}(\Delta_k \mathbf{A})$ and $\overline{\mathbf{B}}_k = (\Delta_k \mathbf{A})^{-1}(\text{exp}(\Delta_k \mathbf{A}) - I)\Delta_k \mathbf{B}_k$. Through this approach, parameters $\overline{\mathbf{A}}_k$, $\overline{\mathbf{B}}_k$, and $\mathbf{C}_k$ become input-dependent. The computational formula for the selective SSM is as follows:
\begin{align}
    &h_k = \overline{\mathbf{A}}_kh_{k-1} + \overline{\mathbf{B}}_kx_k, 
    \label{equ: s6 discrete ssm1}\\
    &y_k = \mathbf{C}_kh_k.
    \label{equ: s6 discrete ssm2}
\end{align}

By correlating model parameters with the input, the selective SSM \ybsun{has the ability to} assign higher parameter weights to key inputs. It thereby prioritizes critical information and filters out irrelevant noisy tokens within the input sequence.
However, this input-dependent parameterization fundamentally alters the computational constraints discussed in Section~\ref{sec: efficient dplr}. \ybsun{In particular, it invalidates} the preconditions for \ybsun{the} efficient computation \ybsun{of} algorithms. Therefore, two \ybsun{recent approaches} have been proposed to improve the efficiency of selective SSMs: the hardware-aware state expansion technique introduced in Mamba and the semiseparable matrices employed in Mamba2. These methods will be discussed in the following two sections.

\subsection{Efficiency: Hardware-aware State Expansion}
\label{sec: Efficient: Hardware-aware State Expansion}

Hardware-aware state expansion is a strategy proposed in Mamba for the efficient computation of Selective SSM. Since model parameters $\overline{\mathbf{A}}$, $\overline{\mathbf{B}}$ and $\mathbf{C}$ are influenced by the input, previous conditions for efficient computations no longer hold. Consequently, Mamba has to revert to the original recurrence method to handle computations associated with SSMs. It optimizes the algorithms related to the underlying hardware to achieve efficient computation.


\ybsun{The optimization of mamba is related to the computational architecture of modern GPUs.} On one hand, Mamba performs the discretization of SSM parameters and recurrence computation directly in the GPU SRAM (Static Random-Access Memory), rather than in the GPU HBM (High-Bandwidth Memory).
SRAM has a higher speed but a smaller memory capacity, whereas HBM offers a larger storage capacity but lower speeds.
Mamba first loads $\mathbf{A}$, $\mathbf{B}$, $\mathbf{C}$, and $\Delta$ from the slow HBM to the fast SRAM. This involves $O(BLD + DN)$ bytes of memory, where $B$, $L$, $D$, and $N$ represent the batch size, sequence length, token embedding dimension, and hidden state dimension. Within the SRAM, it then discretizes $\mathbf{A}$ and $\mathbf{B}$ into $\overline{\mathbf{A}}$ and $\overline{\mathbf{B}}$, as described in Section~\ref{sec: selective}. Next, a parallel associative scan in the SRAM computes the hidden state and ultimately produces the output $y \in \mathbb{R}^{B \times L \times D}$. Finally, the output is written back to the HBM. This algorithm reduces the complexity of IOs from $O(BLDN)$ to $O(BLD)$, resulting in a 20-40 time speedup in practice. Notably, when the input sequence is too long to fit entirely into the SRAM, Mamba splits the sequence and applies the algorithm to each segment, using the hidden state to connect the computations of individual segments.

On the other hand, Mamba \ybsun{also} employs classical recomputation techniques to reduce memory consumption. During the forward pass, Mamba discards the hidden states of size $(B, L, D, N)$ and recomputes them during the backward pass, conserving memory required for storing hidden states. By computing hidden states directly in the SRAM rather than reading them from the HBM, this approach also reduces IOs during the backward pass.


\subsection{Efficiency: Semiseparable Matrices}
\label{sec: Efficient: Semiseparable Matrices}

Mamba2 improves computational efficiency by refining the optimization of the selective SSM formula.
It first reformulates the SSM equation into the form $y = \mathbf{M}x$ and then leverages the properties of $\mathbf{M}$ as a semiseparable matrix to further streamline computations.
This improvement enables Mamba2 to exploit the parallelism and computational optimizations of the Transformer architecture and underlying hardware.

Unlike the iterative computations in Mamba, Mamba2 reformulates the SSM computation (Eq.~\ref{equ: s6 discrete ssm1}) into the form $y = \mathbf{M}x$. Specifically, when setting $h_0 = \overline{\mathbf{B}}_0x_0$, iterating Equation~\ref{equ: s6 discrete ssm1} can yield
\begin{equation}
\begin{aligned}
    h_k &= \overline{\mathbf{A}}_kh_{k-1} + \overline{\mathbf{B}}_kx_k \\
    & = \overline{\mathbf{A}}_k\dots\overline{\mathbf{A}}_1\overline{\mathbf{B}}_0x_0  + \overline{\mathbf{A}}_k\dots\overline{\mathbf{A}}_2\overline{\mathbf{B}}_1x_1 + \dots + \overline{\mathbf{A}}_k\overline{\mathbf{A}}_{k-1}\overline{\mathbf{B}}_{k-2}x_{k-2} + \overline{\mathbf{A}}_k\overline{\mathbf{B}}_{k-1}x_{k-1} + \overline{\mathbf{B}}_{k}x_{k} \\
    & = \sum_{s=0}^{k}\overline{\mathbf{A}}_{k:s}^{\times}\overline{\mathbf{B}}_sx_s, \\
\end{aligned}
\label{equ: mamba2 h}
\end{equation}
where $\overline{\mathbf{A}}_{k:s}^{\times} := \overline{\mathbf{A}}_k\dots\overline{\mathbf{A}}_{(s+1)}$, for $s = k$, $\overline{\mathbf{A}}_{k:k}^{\times} := I$. Substituting the above result into Equation~\ref{equ: s6 discrete ssm2}, 
\begin{equation}
    y_k = \mathbf{C}_kh_k = \sum_{s=0}^{k}\mathbf{C}_k \overline{\mathbf{A}}_{k:s}^{\times}\overline{\mathbf{B}}_sx_s
\end{equation}
can be generated. Equivalently, the relationship between overall output $y$ and input $x$ can be expressed as
\begin{equation}
    y = \mathbf{M}x,
\end{equation}
where $\mathbf{C}_k \in \mathbb{R}^{1 \times N}$ and $\mathbf{M}_{ji}=\mathbf{C}_j \overline{\mathbf{A}}_{j:i}^{\times}\overline{\mathbf{B}}_i$.

$\mathbf{M}$ is the semiseparable matrix, which can be decomposed into several submatrices.
To further optimize computation, Mamba2 uses the blocking and decomposition of matrix $\mathbf{M}$. Specifically, for a large matrix $\mathbf{M} \in \mathbb{R}^{N\times N}$, an appropriate value $n_1$ can be chosen such that $N \mod n_1 =0$. This allows $\mathbf{M}$ to be decomposed into several submatrices $\mathbf{M}_i \in \mathbb{R}^{n_1 \times n_1}$. Equation~\ref{equ: mamba2 M} illustrates this approach for $N=4$ and $n_1=2$. For diagonal submatrices, given that their parameter structure aligns with that of $\mathbf{M}$ but on a smaller scale, the same method can be applied iteratively for further decomposition.
For off-diagonal submatrices, Mamba2 decomposes them into three parts as shown in Equation~\ref{equ: mamba2 M}. This enables efficient matrix multiplication with $x$ using the corresponding algorithm and also allows parallel computation.

\begin{equation}
\begin{aligned}
    \mathbf{M} &= 
    \begin{bNiceArray}{cc|cc}[cell-space-limits=2pt]
    \mathbf{C}_0\overline{\mathbf{A}}_{0:0}^{\times}\overline{\mathbf{B}}_0 & & & \\
    \mathbf{C}_1\overline{\mathbf{A}}_{1:0}^{\times}\overline{\mathbf{B}}_0 & \mathbf{C}_1\overline{\mathbf{A}}_{1:1}^{\times}\overline{\mathbf{B}}_1 & & \\ \hline
    \mathbf{C}_2\overline{\mathbf{A}}_{2:0}^{\times}\overline{\mathbf{B}}_0 & \mathbf{C}_2\overline{\mathbf{A}}_{2:1}^{\times}\overline{\mathbf{B}}_1 & \mathbf{C}_2\overline{\mathbf{A}}_{2:2}^{\times}\overline{\mathbf{B}}_2 & \\
    \mathbf{C}_3\overline{\mathbf{A}}_{3:0}^{\times}\overline{\mathbf{B}}_0 & \mathbf{C}_3\overline{\mathbf{A}}_{3:1}^{\times}\overline{\mathbf{B}}_1 & \mathbf{C}_3\overline{\mathbf{A}}_{3:2}^{\times}\overline{\mathbf{B}}_2 & \mathbf{C}_3\overline{\mathbf{A}}_{3:3}^{\times}\overline{\mathbf{B}}_3 \\
    \end{bNiceArray}
    \\
    & = 
    \begin{bNiceArray}{cc|cc}[cell-space-limits=2pt, columns-width=6.7em]
    \mathbf{C}_0\overline{\mathbf{A}}_{0:0}^{\times}\overline{\mathbf{B}}_0 & & & \\
    \mathbf{C}_1\overline{\mathbf{A}}_{1:0}^{\times}\overline{\mathbf{B}}_0 & \mathbf{C}_1\overline{\mathbf{A}}_{1:1}^{\times}\overline{\mathbf{B}}_1 & & \\ \hline
    \Block{2-2}{
    \begin{bmatrix}
    \mathbf{C}_2\overline{\mathbf{A}}_{2:1}^{\times}\\
    \mathbf{C}_3\overline{\mathbf{A}}_{3:1}^{\times}
    \end{bmatrix}
    \overline{\mathbf{A}}_{1:1}^{\times}
    \begin{bmatrix}
    \overline{\mathbf{A}}_{1:0}^{\times}\overline{\mathbf{B}}_0 &
    \overline{\mathbf{A}}_{1:1}^{\times}\overline{\mathbf{B}}_1
    \end{bmatrix}
    }& & \mathbf{C}_2\overline{\mathbf{A}}_{2:2}^{\times}\overline{\mathbf{B}}_2 & \\
    & & \mathbf{C}_3\overline{\mathbf{A}}_{3:2}^{\times}\overline{\mathbf{B}}_2 & \mathbf{C}_3\overline{\mathbf{A}}_{3:3}^{\times}\overline{\mathbf{B}}_3 \\
    \end{bNiceArray}
    \\
\end{aligned}
\label{equ: mamba2 M}
\end{equation}

These two improvements reduce computational complexity and, more importantly, align the underlying computational logic of selective SSM with that of the Transformer architecture. This allows Mamba2 to benefit from efficient hardware optimizations and parallelization for large-scale training.

\input{sections/4_mamba_sub1}

%% file: sections/4_mamba_sub1.tex
\subsection{Mamba Architecture and Its Variants}
\label{sec: mamba Model Architecture Optimization}

With its intrinsic selectivity and computational efficiency optimizations, the selective SSM offers both effectiveness and efficiency in language-related tasks. Leveraging selective SSM as its core component, Mamba demonstrates robust model capabilities. This section provides a comprehensive overview of Mamba’s architectural framework and key mechanisms, followed by a summary of its various variants. 

Similar to the Transformer, Mamba’s design includes elements such as mapping layers and activation functions. Its key distinction is the integration of the selective SSM module. 
Input data is processed in parallel through two separate pathways, which are then combined via activation or multiplication functions before being passed through a final down-projection layer to produce the output. The first pathway consists of an input up-projection layer, a one-dimensional convolutional layer, an activation function, and the SSM module. The second pathway includes only an input up-projection layer followed by an activation function. 

There are several hybrid architectures and variants of Mamba. Among these, Mamba-transformer models represent a promising approach combining powerful pre-trained Transformers and efficient SSM status. Specifically, the $O(n^2)$ complexity of self-attention in Transformer models leads to high memory usage for long sequence modeling,
\ybsun{while }the SSM architecture design bypasses this issue. Jamba~\citep{lieber2024jamba} proposed a hybrid model that integrates vanilla Transformer layers, Mamba layers, and MoE layers, achieving a context length of 256K tokens on a single 80GB GPU. Xu et al.~\citep{Xu2024SSTMH} employed Mamba to capture broad trends in long-range sequences and Local Window Transformer for finer details in short-range data. Furthermore, Wang et al.~\citep{wang2024mamba} distilled pre-trained transformers, such as Llama, into a linear RNN by projecting self-attention weights. This allows powerful LLMs to serve as the initial parameters for SSM states.

Besides integrating Mamba with Transformers, recent works have also optimized Mamba internally, adapting it to specific task requirements. Behrouz et al.~\citep{behrouz2024mambamixer} introduce a weighted averaging mechanism to connect selective mixers, allowing layers to directly attend to early-stage features. Pi’oro et al.~\citep{Pioro2024MoEMambaES} and Anthony et al.~\citep{Anthony2024BlackMambaMO} combine SSMs with MoEs to improve scalability in sequential modeling, achieving improvements in inference and training FLOPs. 
DenseSSM~\citep{He2024DenseMambaSS} injects shallow-layer states into deeper layers, thus preserving fine-grained details while maintaining training parallelizability and inference efficiency.
HiSS~\citep{Bhirangi2024HierarchicalSS} constructs a temporal hierarchy by stacking structured SSMs for continuous sequential prediction, which significantly improves performance over state-of-the-art sequence models. Ahamed and Cheng~\citep{Ahamed2024MambaTabAP} fine-tune structured SSMs specifically for tabular data. Additionally, Wang et al.~\citep{Wang2024MambaByteTS} introduce MambaByte, a model trained directly on byte sequences for language modeling. By incorporating speculative decoding, MambaByte achieved a $2.6\times$ inference speedup. 

\subsection{Analyses of Selective State Space Models}

\label{sec: mamba Theoretical Analysis}

This section describes the use of theoretical and practical tools to analyze the selective SSM and gain further insights. Analysis focuses on in-context learning (ICL), since the long-context memorization and generalization abilities that arise from the SSM structure significantly impacts results.
Using tools from Rough Path Theory, Cirone et al.~\citep{Cirone2024TheoreticalFO} demonstrate that the hidden state can capture non-linear interactions across timescales, i.e. a low-dimensional projection of the input’s signature. This theory explains the superior accuracy and efficiency of selective SSMs.
Park et al.~\citep{pmlr-v235-park24j} evaluate the ICL capabilities of SSMs, revealing that selective SSMs achieve comparable performance to Transformers on standard regression tasks and excel in sparse parity learning. Based on these findings, they propose hybrid architectures as a promising approach to further enhance ICL performance. Jelassi et al.~\citep{Jelassi2024RepeatAM} demonstrate that a 2-layer transformer can copy strings of exponential length, whereas SSMs are constrained by their fixed-size latent states. Empirically, in string copying tasks, Transformers outperformed SSMs in both efficiency and generalization. Additionally, Akyürek et al.~\citep{Akyrek2024InContextLL} identify "n-gram heads" as the key factor enabling Transformers to outperform SSMs in in-context learning. They further demonstrate that incorporating these heads into selective SSMs improves performance in natural language modeling.

%% file: sections/5_models.tex
One compelling aspect of the SSM is its deep connections with other model architectures. The original equations defining SSMs naturally correspond to the structure of recurrent neural networks (RNNs)~\citep{elman1990finding}. With straightforward derivations, they can also align with convolutional neural networks (CNNs)~\citep{lecun1998gradient}.
Several works have been inspired by the combinations of SSMs, RNNs, and CNNs~\citep{sun2024learning, bick2024transformers, hwang2024hydra,bick2025llamba}.
Furthermore, as SSM models evolve, their formulations and practical implementations increasingly converge with Transformer models~\citep{vaswani2017attention,dao2024transformers}. Sections~\ref{SSM and RNN, CNN},~\ref{SSM and Transformer} explore these relationships, focusing on the mathematical links and developmental trajectories of SSM in relation to RNN, CNN, and Transformer.

\subsection{SSM and RNN, CNN}
\label{SSM and RNN, CNN}

Mathematically, SSM has strong connections to RNN and CNN. Although the SSM series was initially introduced as an independent framework, its original equations align with the simplified structure of RNN. 
In the discrete-time SSM equations (Eq.~\ref{equ: discrete ssm1},\ref{equ: discrete ssm2}), parameters $\overline{\mathbf{A}}$ and $\overline{\mathbf{B}}$ correspond to the hidden layer parameters of an RNN, $\mathbf{C}$ and $\mathbf{D}$ correspond to its output layer parameters, and $h_k$ represents the hidden states. In the time-invariant scenarios where SSM parameters remain invariant to input, these equations can be reformulated as a convolutional operation (Eq.~\ref{equ: conv with length L}), which constitutes the core computation in CNN.

This mathematical correspondence elucidates the strengths and weaknesses of SSMs. For example, while SSMs can reduce computational complexity, they may suffer from memory decay similar to RNNs. Such insights support theoretical analysis and architectural improvements. The mathematical flexibility of SSMs also opens up numerous optimization opportunities. A prime example is the S4 model, which uses the convolutional equivalence of SSM to significantly enhance computational efficiency~\citep{gu2021efficiently}.

\subsection{SSM and Transformer}
\label{SSM and Transformer}

Throughout their development, SSM models have converged toward the Transformer architecture. Mamba~\citep{gu2023mamba}, Mamba2~\citep{dao2024transformers}, and subsequent research have focused on optimizing SSM formulations to take advantage of hardware acceleration and parallelism, similar to Transformers. For example, Mamba2 reformulates SSM computations into the matrix-vector multiplication format 
$y = \mathbf{M}x$, aligning with efficient computational frameworks. Moreover, starting from S4, the design of SSM-based models has increasingly mirrored that of Transformers, with attention modules being replaced by SSM modules, along with other minor modifications~\citep{fathi2023block-state,zuo2022efficient,dao2024transformers}.
There has also been growing interest in connections between SSM and attention mechanisms. Notably, the equivalence between SSM and linear attention has already been established~\citep{dao2024transformers}. Recent works have further proposed fusing the two architectures by directly integrating SSM-based model blocks into Transformers~\citep{lieber2024jamba}.

This trend is primarily driven by practical application demands. Early SSMs, such as S4, showcased computational efficiency and the ability to handle long sequences, but their real-world applicability was limited due to their poor compatibility with modern hardware and subpar performance on practical tasks. In contrast, Transformer models excel in both hardware utilization and task performance. The shift toward Transformer-like structures has improved the performance and applicability of SSM models, broadening their adoption in practical scenarios.

%% file: sections/10_application.tex
After covering the theoretical foundations of the original SSM, S4, and Mamba, we introduce their applications in downstream tasks and data contexts.
A pivotal question arises before applying SSM-based models to practical tasks: \textit{``What is the most significant characteristic of SSM-based models in downstream applications?''}
First, state space models are highly efficient at modeling long-range dependencies, which enables them to excel in sequence data modeling. This includes text represented as token sequences, videos as numerical frames, and molecular sequences like DNA, RNA, and proteins, among other data categories.
Second, enhanced sequence modeling capabilities significantly extend the receptive field of SSM-based models, facilitating their application in diverse image-related and specialized tasks.



\input{figures/application}


\subsection{Video Modeling}
\label{sec:application_video}
SSMs enable efficient and scalable video modeling, improving performance across a wide array of video understanding and generation tasks. Numerous recent works integrate SSM modules—often through variants of the Mamba operator—into different video architectures to better capture long-term dependencies while maintaining computational efficiency. For instance, Vivim~\citep{yang2024vivim} and VideoMamba~\citep{li2025videomamba} use SSM-inspired operations for medical video segmentation and general video action recognition, respectively. This balances broad temporal context with computational tractability. Similarly, SpikeMba~\citep{li2024spikemba} employs SSMs alongside spiking neural networks to refine temporal video grounding and reduce confidence biases. RainMamba~\citep{wu2024rainmamba} enhances video deraining by improving locality modeling through a Hilbert scanning mechanism. Other lines of research, such as RhythmMamba~\citep{zou2024rhythmmamba} and Simba~\citep{chaudhuri2024simba}, use SSM blocks to efficiently model periodic physiological signals for remote photoplethysmography and to strengthen skeleton-based action recognition with structural priors and temporal reasoning. Mamba4D~\citep{liu2024mamba4d} integrates SSM-based modules into point cloud video analysis for effective long-term spatiotemporal modeling in complex 4D data. For generative tasks, Matten~\citep{gao2024matten} and Diffusion Mamba~\citep{mo2024scaling} adopt SSM operators to achieve efficient, scalable video synthesis and latent diffusion without the quadratic complexity of self-attention. DeMamba~\citep{chen2024demamba} applies a similar principle for detecting AI-generated videos at scale. These studies highlight the versatility and effectiveness of SSM frameworks for video understanding and generation.

\subsection{Speech and Audio Modeling}
\label{sec:application_speech}


In diverse speech and audio processing tasks, SSMs, notably Mamba and its variants, have demonstrated increasing potential. In speech enhancement, Mamba-based networks effectively capture long-term temporal dependencies and show robust performance in both static and moving speaker scenarios. A representative example is the streaming SpatialNet variant, where Mamba replaces self-attention modules to achieve linear-time inference and strong generalization on extended audio streams~\citep{quan2024multichannel}. Similarly, Audio Mamba (AuM)~\citep{erol2024audio} demonstrates that a fully SSM-based architecture can match or exceed the performance of established Audio Spectrogram Transformers without relying on self-attention. Beyond enhancement and classification, Mamba offers a scalable alternative to Transformers for speech recognition, as shown by the improved semantic-aware modeling of Bidirectional Mamba (BiMamba) compared to its vanilla counterpart~\citep{zhang2024mamba}. In self-supervised learning, Ssamba~\citep{shams2024ssamba} uses Mamba to capture rich contextual information in unlabeled audio, with better efficiency and accuracy than self-attention-based models. For speech enhancement tasks specifically, SEMamba~\citep{chao2024investigation} incorporates Mamba’s efficient state-space formulation into regression-based architectures, achieving state-of-the-art perceptual evaluation scores. In speech separation, both SPMamba~\citep{li2024spmamba} and Dual-path Mamba~\citep{jiang2024dual} exploit the linear complexity and bidirectionality of SSMs to surpass conventional recurrent or attention-based systems in handling long audio sequences, ultimately improving separation quality and computational efficiency. Evidently, Mamba-based SSMs are scalable, effective, and memory-efficient alternatives to Transformer-driven architectures for speech and audio applications.

\subsection{Molecular Modeling}
\label{sec:application_molecular}

SSMs are also contributing to molecular modeling tasks within biology and chemistry. In genomics, long-range modeling capabilities are essential for understanding the context-dependent functions of regulatory elements within DNA sequences. Caduceus~\citep{schiff2024caduceus} integrates a bi-directional Mamba component (BiMamba) and reverse complement (RC) equivariance to improve variant effect prediction, outperforming larger non-equivariant models. Likewise, HyenaDNA~\citep{nguyen2024hyenadna} implements long-range sequence modeling without attention, enabling single nucleotide-level resolution across up to one million tokens and achieving state-of-the-art performance on a range of genomic benchmarks. Beyond genomics, SSM-based architectures show promise in chemical informatics. By pre-training Mamba-based models on massive SMILES corpora, researchers have developed efficient and scalable chemical foundation models that deliver state-of-the-art results in property prediction, classification, and synthesis yield prediction~\citep{brazil2024mamba}. SMILES-Mamba~\citep{xu2024smiles} employs self-supervised pretraining followed by fine-tuning to substantially enhance drug ADMET prediction accuracy while mitigating the reliance on large labeled datasets. Finally, these successes extend to biomedical text representation. BioMamba~\citep{yue2024biomamba} efficiently parses and interprets complex biomedical literature, surpassing the performance of domain-specific Transformers. Collectively, these studies show how the advantages of SSMs in capturing long-range dependencies, contextual nuances, and structural patterns support crucial improvements in molecular modeling.

\subsection{3D Signals Modeling}
\label{sec:application_3d_signal}
We next discuss the applications of SSMs in modeling complex 3D signals, spanning tasks from single-view reconstruction and medical imaging to point cloud analysis. For 3D asset generation, Gamba~\citep{shen2024gamba} relies on a Mamba-based sequential prediction backbone and Gaussian Splatting. It achieves end-to-end single-image 3D reconstruction with millisecond-level speed, significantly outperforming optimization-based counterparts. In medical applications, SegMamba~\citep{xing2024segmamba} demonstrates that Mamba-based architectures can efficiently handle large volumetric data and model long-range dependencies for 3D medical image segmentation. Similarly, nnMamba~\citep{gong2024nnmamba} merges CNNs with SSMs for 3D segmentation, classification, and landmark detection, achieving state-of-the-art performance across biomedical imaging benchmarks. Complementing these efforts, CMViM~\citep{yang2024cmvim} integrates Vision Mamba (Vim) into a masked autoencoding framework for 3D multi-modal data representation, enabling more accurate Alzheimer’s disease classification. Other than voxelized representations, SSM-based models such as PointMamba~\citep{liang2024pointmamba} and Mamba3D~\citep{han2024mamba3d} facilitate point cloud tasks through their linear-complexity global modeling ability. PointMamba significantly reduces computational costs while maintaining effective global feature extraction, whereas Mamba3D enhances local geometric and global contextual modeling, surpassing Transformer-based methods on standard benchmarks. Furthermore, 3DMambaIPF~\citep{zhou20243dmambaipf} employs selective SSMs for large-scale point cloud denoising and incorporates differentiable rendering to refine geometric accuracy on massive datasets. It can be seen that SSM-based architectures can efficiently capture long-range dependencies and complex structures across a wide range of 3D signals and domains.

\subsection{Time Series Modeling}
\label{sec:application_time_series}

For time series modeling, SSMs have similarly emerged as a compelling approach to challenges such as long-range modeling and efficiency. Several works have demonstrated the versatility of Mamba-based SSMs across time series forecasting tasks. For instance, by integrating Vision Mamba blocks into recurrent architectures, VMRNN~\citep{tang2024vmrnn} combines the strengths of LSTMs with the linear complexity of Mamba for improved spatiotemporal prediction. Moving toward longer-term prediction, TimeMachine~\citep{ahamed2024timemachine} leverages multiple Mamba modules to effectively model both channel-mixing and channel-independent patterns, improving the scalability and memory efficiency of long-range forecasting. S-Mamba~\citep{wang2024mambab} shows that, compared to Transformers, Mamba can reduce complexity while retaining superior predictive performance, which benefits real-world deployment. Beyond these single-model paradigms, SiMBA~\citep{patro2024simba} merges Mamba with spectral techniques (EinFFT) for channel modeling, narrowing the gap between SSMs and Transformers on both vision and multivariate time series benchmarks. Hybrid solutions have also surfaced: MAT~\citep{xu2024integrating} integrates Mamba and Transformer components to capture both long- and short-range dependencies, improving accuracy and resource utilization on weather prediction tasks. Bi-Mamba+~\citep{liang2024bi} introduces a gating mechanism and bidirectional processing to refine Mamba’s contextual understanding. It represents a more flexible, data-driven approach that adapts to diverse time series structures.

\subsection{Structured Data Modeling}
\label{sec:application_structured}


Finally, we cover the applications of SSMs in structured data modeling. In graph representation learning, Graph Mamba Networks (GMNs)~\citep{behrouz2024graph} use selective SSMs to accurately and efficiently capture long-range dependencies in graphs. This approach alleviates over-squashing and maintains strong performance on small- and large-scale benchmarks without complex positional encodings. Similarly, STG-Mamba~\citep{li2024stg} extends the selective SSM paradigm to spatial-temporal graph (STG) data, addressing the intricate dynamic evolution and heterogeneity in such systems. It integrates a Spatial-Temporal Selective State Space Module (ST-S3M) and Kalman Filtering-inspired GNN layers to achieve state-of-the-art forecasting results at reduced computational costs. In the tabular domain, MambaTab~\citep{ahamed2024mambatab} adapts Mamba-based architectures through a lightweight ``plug-and-play'' approach that bypasses extensive preprocessing steps and parameter tuning. Using fewer parameters, MambaTab outperforms established baselines on a range of tabular datasets. The above works on structured data modeling span graphs, time-evolving networks, and tabular structures, showcasing the adaptability of SSM-based models.

%% file: figures/application.tex
\begin{figure*}
\footnotesize
        \begin{forest}
            for tree={
                forked edges,
                grow'=0,
                draw,
                rounded corners,
                node options={align=center,},
                text width=3.5cm,
                s sep=6pt,
                calign=child edge, calign child=(n_children()+1)/2,
            },
            [Application, fill=gray!45, parent
                [Video Modeling \S\ref{sec:application_video}, for tree={o_ssm}
                    [   Video Understanding, o_ssm
                        [VideoMamba~\citep{li2025videomamba}; Vivim~\citep{yang2024vivim}; Spikemba~\citep{li2024spikemba}; Rainmamba~\citep{wu2024rainmamba}; Rhythmmanba~\citep{zou2024rhythmmamba}; Simba~\citep{chaudhuri2024simba}; Mamba4D~\citep{liu2024mamba4d}, o_ssm_work]
                    ]
                    [
                        Video Generation, o_ssm
                        [Matten~\citep{gao2024matten}; DeMamba~\citep{chen2024demamba}; DiM~\citep{mo2024scaling}, o_ssm_work]
                    ]
                ]
                [Speech and Audio \\Modeling \S\ref{sec:application_speech}, for tree={s4}
                    [Speech Enhancement,  s4
                        [SEMamba~\citep{chao2024investigation}; SpatialNet~\citep{quan2024multichannel}; Spmamba~\citep{li2024spmamba}; Dual-path Mamba~\citep{jiang2024dual}, s4_work]
                    ]
                    [Speech Representation,  s4
                        [Audio Mamba~\citep{erol2024audio}; MambaSpeech~\citep{zhang2024mamba}; Ssamba~\citep{shams2024ssamba}, s4_work]
                    ]
                ]
                [Molecular \\ Modeling \S\ref{sec:application_molecular}, for tree={mamba}
                    [Biology and Chemistry, mamba
                        [Hyenadna~\citep{nguyen2024hyenadna};Caduceus~\citep{schiff2024caduceus};ChemMamba~\citep{brazil2024mamba}; Smiles-mamba~\citep{xu2024smiles};Biomamba~\citep{yue2024biomamba}, mamba_app]
                    ]
                ]
                [3D Signals \\ Modeling \S\ref{sec:application_3d_signal}, for tree={o_ssm}
                    [3D Representation, o_ssm
                        [Gamba~\citep{shen2024gamba};Segmamba~\citep{xing2024segmamba};nmamba~\citep{gong2024nnmamba};CMViM~\citep{yang2024cmvim} , o_ssm_work]
                    ]
                    [Point Cloud, o_ssm
                        [Pointmamba~\citep{liang2024pointmamba};Mamba3d~\citep{han2024mamba3d}; 3dmambaipf~\citep{zhou20243dmambaipf}, o_ssm_work]
                    ]
                ]
                [Time Series \\ Modeling \S\ref{sec:application_time_series}, for tree={s4}
                    [General Time Series, s4
                        [Timemachine~\citep{ahamed2024timemachine};TSF-Mamba~\citep{wang2024mambab};MAT;Bio-Mamba4TS~\citep{liang2024bi} , s4_work]
                    ]
                    [Spatiotemporal, s4
                        [Vmrnn~\citep{tang2024vmrnn};Simba~\citep{patro2024simba} , s4_work]
                    ]
                ]
                [Structured Data \\ Modeling \S\ref{sec:application_structured}, for tree={mamba}
                    [Graph and Table, mamba
                        [Graph Mamba~\citep{behrouz2024graph}; stg-mamba~\citep{li2024stg}; Mambatab~\citep{ahamed2024mambatab}, mamba_app]
                    ]
                ]
            ]
        \end{forest}
            \caption{Applications of State Space Models Across Diverse Data Types}
            \label{fig:typo-prompt}
\end{figure*}
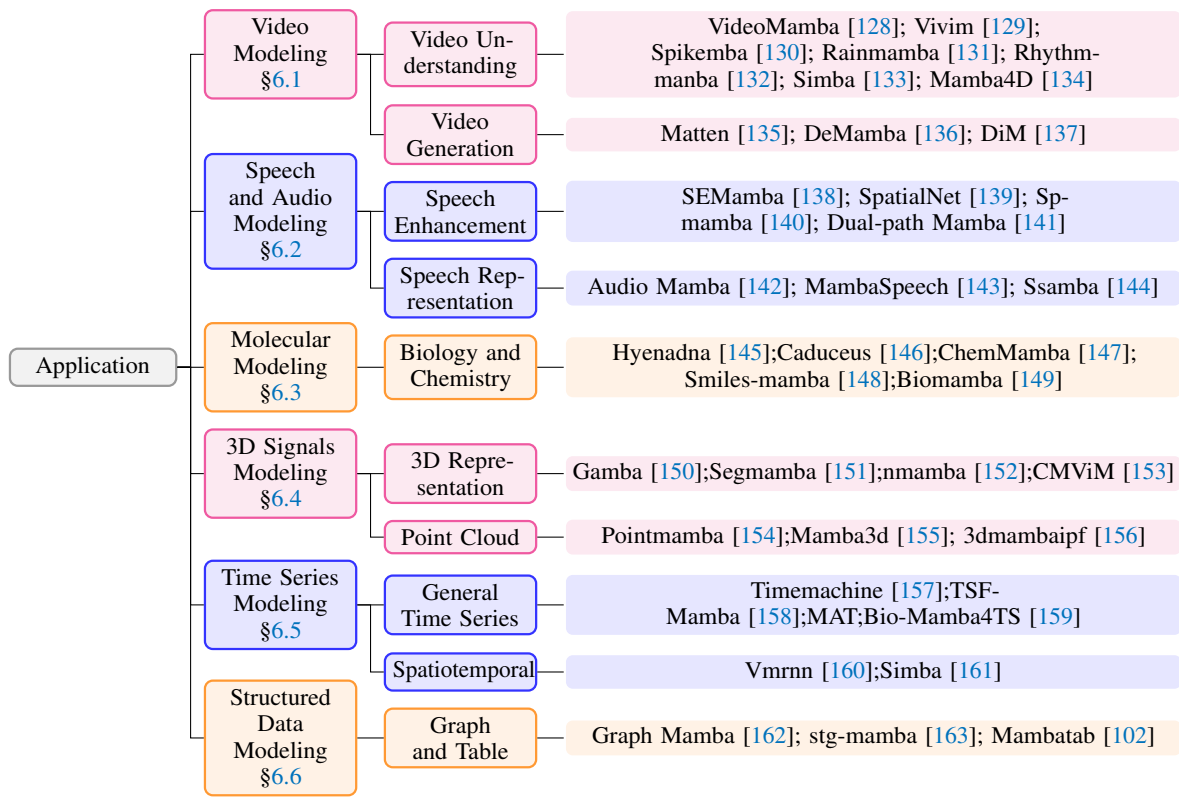